# From Landslide Conditioning Factors to Satellite Embeddings: Evaluating the Utilisation of Google AlphaEarth for Landslide Susceptibility Mapping using Deep Learning

Yusen Cheng[a, b], Qinfeng Zhu[a, c], Lei Fan[a, *]

[a] Department of Civil Engineering, Xi'an Jiaotong-Liverpool University, Suzhou, 215123, China

[b] Department of Geography and Planning, University of Liverpool, Liverpool, L69 3BX, UK

[c] Department of Computer Science, University of Liverpool, Liverpool, L69 3BX, UK

[*] Corresponding author (Email: Lei.Fan@xjtlu.edu.cn)

## Highlights

- First use of AlphaEarth embeddings for landslide susceptibility mapping
- AlphaEarth outperforms conventional factors across regions and models
- AlphaEarth improves spatial discrimination of landslide-prone areas
- AlphaEarth is validated as effective input for landslide susceptibility mapping

## Abstract

Data-driven landslide susceptibility mapping (LSM) typically relies on landslide conditioning factors (LCFs), whose availability, heterogeneity, and preprocessing-related uncertainties can constrain mapping reliability. Recently, Google AlphaEarth (AE) embeddings, derived from multi-source geospatial observations, have emerged as a unified representation of Earth surface conditions. This study evaluated the potential of AE embeddings as alternative predictors for LSM. Two AE representations, including retained principal components and the full set of 64 embedding bands, were systematically compared with conventional LCFs across three study areas (Nantou County, Taiwan; Hong Kong; and part of Emilia-Romagna, Italy) using three deep learning models (CNN1D, CNN2D, and Vision Transformer). Performance was assessed using multiple evaluation metrics, ROC-AUC analysis, error statistics, and spatial pattern assessment. Results showed that AE-based models consistently outperformed LCFs across all regions and models, yielding higher F1-scores, AUC values, and more stable error distributions. Such improvement was most pronounced when using the full 64-band AE representation, with F1-score improvements of approximately 4% to 15% and AUC increased ranging from 0.04 to 0.11, depending on the study area and model. AE-based susceptibility maps also exhibited clearer spatial correspondence with observed landslide occurrences and enhanced sensitivity to localised landslide-prone conditions. Performance improvements were more evident in Nantou and Emilia than in Hong Kong. SHAP analysis further revealed regional variability in the contribution of individual AE bands, while several bands showed consistently high importance across all three study areas. These findings highlight the strong potential of AE embeddings as a standardised and information-rich alternative to conventional LCFs for LSM.

# 1 Introduction

Landslides are among the most damaging natural hazards worldwide and can be triggered by a wide range of factors, including earthquakes, rainfall, snowmelt, volcanic activity, and human interventions (Hungr et al., 2014). They often lead to substantial loss of life, severe damage to transportation networks and critical infrastructure, and long-term socioeconomic disruption (Raška et al., 2023; Joshi et al., 2025). In recent decades, the frequency and impact of landslide disasters have shown an increasing trend, primarily driven by climate change and associated shifts in extreme weather patterns (Jemec Auflič et al., 2023; Liu & Wang, 2024; C.-W. Chen et al., 2025). According to the 2024 Global Natural Disaster Assessment Report, landslides accounted for approximately 10.68% of all recorded natural hazards worldwide in 2024, accompanied by a notable rise in both fatalities and affected populations (BNU et al., 2025).

Landslide susceptibility mapping (LSM) aims to estimate the spatial likelihood of landslide occurrence under given environmental conditions (Guzzetti et al., 2005). In recent years, quantitative approaches have become dominant in LSM research, broadly categorised into two main types: physically based models and data-driven methods. Physically based models are grounded in slope stability theory and hydrological processes, allowing them to represent the mechanisms associated with landslide initiation (Medina et al., 2021; Durmaz et al., 2023). As a result, they are handy for site-scale analyses and engineering-oriented applications. Nevertheless, their extension to regional-scale LSM remains challenging, mainly due to the limited availability of reliable geotechnical and hydrological parameters over large areas, as well as the substantial computational effort required to simulate complex processes (Marin et al., 2021; Rana & Babu, 2022; Li et al., 2025). To address these constraints, increasing efforts have been made to estimate landslide-related physical parameters using more readily available environmental and soil information. For example, soil thickness has been inferred from empirical relationships with Digital Elevation Model (DEM) derived terrain attributes, including elevation, slope, and hillslope curvature (Patton et al., 2018; Fusco et al., 2021). Similarly, effective cohesion has been spatially represented using vegetation information, with Normalised Difference Vegetation Index (NDVI) serving as a proxy for the contribution of root reinforcement to soil strength (Jr-Chuan Huang et al., 2006). Residual friction angle has also been estimated indirectly from soil composition (Pei & Qiu, 2024). Beyond such parameter-specific empirical relationships, increasingly available gridded soil datasets and predictive tools provide a more systematic basis for regional parameterisation. Global products such as SoilGrids provide machine-learning-based estimates of soil texture, bulk density, depth to bedrock, and other soil properties (Hengl et al., 2017), while regional studies have combined machine learning with spatial interpolation to construct distributed databases of saturated and unsaturated hydraulic and mechanical properties (Li et al., 2022). Pedotransfer approaches (PTFs) such as ROSETTA can additionally estimate van

Genuchten water-retention parameters and saturated hydraulic conductivity from relatively accessible soil information, ranging from textural class to more detailed inputs such as particle-size fractions and bulk density (Schaap et al., 2001). Although these approaches can reduce reliance on dense field sampling and facilitate the regional application of physically based models, the derived parameters may remain considerable uncertainty.

In contrast, data-driven models offer greater flexibility for large-scale applications by learning underlying relationships between historical landslide occurrences and a set of landslide conditioning factors (LCFs). These factors commonly include terrain attributes (e.g., elevation, slope, and curvature), hydrological variables (e.g., rainfall), vegetation-related proxies (e.g., NDVI), land use/land cover, and geological information. Within this paradigm, the selection and representation of LCFs play a critical role in shaping LSM results. Previous studies have examined how LCFs influence the performance of data-driven LSM. For example, Chen and Fan (2023) compared four different LCF selection strategies and reported that model performance generally increased with the number of included factors up to a certain threshold, beyond which performance began to decline. Similar trends have also been observed in other studies (Gu et al., 2023; Wang et al., 2025), indicating that optimising LCF combinations can enhance susceptibility prediction, while also increasing workflow complexity (Huang et al., 2025). Another potential limitation of LCF-based LSM lies in its reliance on hand-crafted indices derived from raw observations. A typical example is land use/land cover (LULC), which compresses multispectral information into discrete thematic classes. The generation of LULC maps typically depends on classification algorithms and, in many cases, subjective decisions related to class definitions, which may introduce additional uncertainties and propagate errors into subsequent LSM analyses (Yu et al., 2023). Alternatively, remote sensing imagery can be used more directly in landslide-related modelling without first converting the spectral information into predefined thematic factors. Previous studies have employed multispectral satellite imagery with machine learning, unsupervised deep learning, and more recently self-supervised learning for landslide detection and mapping (Tavakkoli Piralilou et al., 2019; Shahabi et al., 2021, 2026). More generally, uncertainties arising from LCF construction, measurement errors, and data preprocessing have been shown to substantially affect LSM outcomes, even when the modelling framework itself remains unchanged (Huang, Teng, et al., 2024).

Beyond these challenges, inconsistencies in spatial resolution across different LCFs pose an additional difficulty for practical LSM applications. For instance, LCFs derived from DEMs or optical imagery often have relatively fine spatial resolutions (e.g., 30 m or finer), whereas rainfall information derived from either satellite products or rain-gauge networks is typically provided at much coarser resolutions, often several kilometres. Resampling or interpolating such heterogeneous datasets to a standard grid can introduce additional uncertainties and smoothing effects (Salleh et al., 2024), which may influence model performance in ways that are difficult to quantify. In addition, from a data availability perspective, geological factors such as lithology and fault lines are also frequently included in LSM due to their relevance to landslide processes (Reichenbach et al., 2018; Cheng et al., 2025). However, these datasets commonly rely on field-based geological surveys and are not consistently available across regions. Even where available, they often vary in update frequency and level of detail.

The recent release of Google AlphaEarth (AE) embeddings, generated from Google DeepMind's AlphaEarth Foundations (AEF) model, has begun to reshape how environmental information is represented and used across a wide range of geospatial applications (Brown et al., 2025). Rather than being designed for a specific hazard type or, more broadly, a single geoscientific task, the AEF model aims to learn general-purpose representations of the Earth's surface and near-surface environment from large-scale, multi-source geospatial observations, including optical imagery, SAR, LiDAR, meteorological variables, and GRACE-based hydrological information. By compressing complex surface characteristics into a group of unified embeddings at a consistent spatial resolution, AE embeddings provide a compact and information-rich representation that can be readily integrated into data-driven geoscientific modelling workflows. Since their release, AE embeddings have demonstrated strong potential across diverse geoscientific tasks, including crop mapping (Murakami, 2025), urban air quality prediction (Alvarez et al., 2025), and groundwater fluoride prediction (Wei et al., 2025). From the perspective of LSM, AE embeddings also show clear potential. Compared to LCF-based inputs, AE provides a standardised and unified set of predictors, thereby reducing the need for extensive preprocessing steps, such as factor selection, resampling, and reclassification, which are common sources of uncertainty in LSM workflows. In addition, as AE embeddings are learned directly from multi-source observations rather than derived indices, they may preserve more informative environmental characteristics than LCFs. This richer representation may help capture subtle spatial patterns and contextual signals that are relevant to landslide occurrence but difficult to express using a set of LCFs.

Against this background, this study explores the potential of AE embeddings as alternative predictors for data-driven LSM. AE-derived predictors are systematically compared with conventional LCFs across three study areas and three deep learning models, to assess whether AE can deliver more reliable LSM results and under which conditions their advantages are most evident. The contribution of this study is summarised as follows:

- To the best of our knowledge, this study represents the first attempt to employ Google AE embeddings to landslide research, with a specific focus on LSM.
- The differences in spatial patterns of landslide susceptibility maps produced using AE-based predictors and conventional LCF-based inputs are investigated and analysed.
- The superior predictive performance of AE-derived predictors relative to LCFs is demonstrated across

multiple study areas and deep learning models.

- The study offers new insights into the mechanisms governing AE-based LSM performance, identifying key factors that influence model effectiveness.

# 2 Study Area and Data

## 2.1 Study Areas and Samples

To systematically evaluate the practical utility of AE embeddings in LSM, three study areas with distinct environmental and geomorphological settings were selected: Nantou County in central Taiwan, Hong Kong, and the Brisighella and Modigliana communes in the Emilia-Romagna region of Italy (Fig. 1). Nantou County, located in central Taiwan, exhibits pronounced topographic relief, with elevations decreasing rapidly from over 3800 m in the eastern mountainous part of the county to less than 100 m within a horizontal distance of several tens of kilometres. Hong Kong represents a highly urbanised coastal region, where steep slopes coexist with dense infrastructure and intense human activities. The Emilia study area is characterised by hilly to mountainous terrain with relatively continuous slopes and moderate topographic relief.

Reliable landslide inventory datasets are a fundamental prerequisite for landslide susceptibility studies (Huang, Li, et al., 2024). In this study, landslide inventories for the three study areas were obtained from authoritative sources or previous research (Table 1). For Nantou County, the landslide inventory was derived from the 2022 annual Taiwan landslide inventory. This annually updated polygon inventory was treated as a snapshot of landslide conditions in 2022, and landslide samples were represented by points extracted from the crown areas of mapped landslides. In Hong Kong, landslide data were obtained from the Enhanced Natural Terrain Landslide Inventory (ENTLI), which is already a point-based inventory. As the inventory is progressively updated by adding newly identified landslides, only records from 2000 to 2019 were retained. This cumulative structure introduces some temporal uncertainty as earlier mapped landslides may have subsequently been altered, but such changes cannot be reliably reconstructed from the available inventory. For the Emilia-Romagna study area, the inventory was derived from Ferrario and Livio (2023), which mapped landslides triggered by two major rainfall events in May 2023. These post-event landslide polygons were likewise represented by points extracted from their crown areas. As all three inventories are vector-based, all landslide and non-landslide samples were ultimately aligned to a unified 30 m grid for model construction. Potential positional uncertainty associated with the original landslide inventory was retained, and no additional relocation or spatial correction was applied. Exact event dates were not used as model inputs, with the inventories serving to construct static susceptibility datasets representative of the corresponding study periods. In addition, non-landslide samples were randomly selected outside a 150 m buffer surrounding the recorded landslides to reduce potential labelling uncertainties (Zhu et al., 2024). A 1:1 ratio between landslide and non-landslide samples was adopted for each study area. Thus, balanced datasets were used for modelling, and no additional class-imbalance treatment was required.

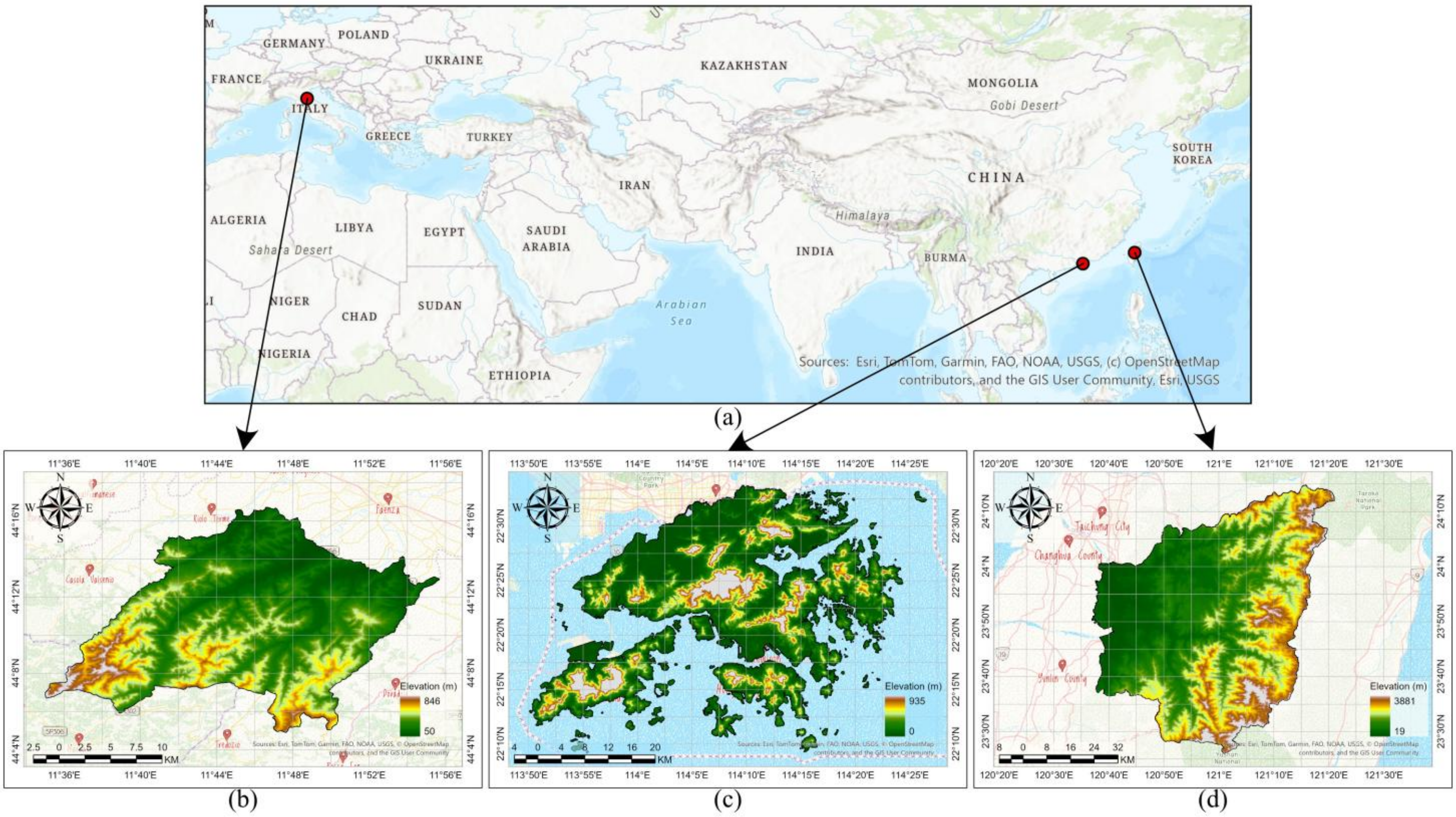

**Fig. 1**. Locations of the study areas: **(a)** Geographical locations of the three study areas; **(b)** Brisighella and Modigliana communes, Emilia-Romagna region, Italy; **(c)** Hong Kong; **(d)** Nantou County, Taiwan.

**Table 1**. Summary of landslide inventory datasets used in this study.

| Inventory Information | Nantou County, Taiwan | Hong Kong | Brisighella and Modigliana communes, Emilia-Romagna region, Italy |
|---|---|---|---|
| Number of Landslides | 6075 | 7542 | 2614 |
| Inventory Period | 2022 | 2000-2019 | 2023 |
| Inventory Geometry | Polygon | Point | Polygon |
| Temporal Representation | Annual snapshot | Cumulative inventory | Post-event inventory |
| Data Source | 111 Annual Taiwan Landslide Inventory | Hong Kong ENTLI Dataset | (Ferrario & Livio, 2023) |
| URL | https://data.gov.tw/en/datasets/166846 | https://www.ginfo.cedd.gov.hk/geoopendata/eng/ENTLI.aspx | https://zenodo.org/records/8102429 |

## 2.2 Landslide Conditioning Factors

In this study, fourteen LCFs commonly used in previous landslide susceptibility studies were selected to characterise key environmental conditions associated with landslide occurrence across the three study areas (Cheng et al., 2025). These factors were grouped into five categories: topographic, geological, hydrological, surface environmental, and human-related factors.

The topographic category encompasses elevation-derived variables that describe terrain morphology and surface complexity, including elevation, slope, aspect, curvature, and the topographic roughness index (TRI). Additionally, it includes hydrologically relevant terrain indices, such as the stream power index (SPI) and topographic wetness index (TWI). Geological factors consist of lithology and distance to faults, representing variations in geological materials and structural features. Hydrological conditions are represented by maximum daily rainfall (RX1D), which was used to characterise landslide susceptibility under extreme rainfall scenarios. For Nantou and Emilia, the 8-km CMORPH rainfall data were bilinearly resampled to 30 m, whereas for Hong Kong, RX1D was derived from observations at 90 rain gauges and spatially interpolated using kriging. Surface environmental conditions are characterised using NDVI and land use and land cover (LULC), which reflect differences in vegetation cover and surface properties. Human-related influence is captured through distance to roads, accounting for terrain modification and drainage alteration associated with infrastructure development.

All LCFs were prepared as raster layers and resampled to a uniform spatial resolution of 30 m × 30 m to ensure spatial consistency across datasets and study areas. Detailed information on data sources for each LCF is summarised in Table 2.

**Table 2**. LCFs and corresponding data sources for the three study areas.

| **Data** | **Study areas** | | | **LCFs** |
|---|---|---|---|---|
| | **Hong Kong** | **Nantou** | **Emilia** | |
| DEM | ASTER Global DEM V003 (30m) (NASA/METI/AIST/Japan Spacesystems and U.S./Japan ASTER Science Team, 2019) | | | Elevation, slope, aspect, curvature, SPI, TWI, and TRI. |
| Lithology | Common Spatial Data Infrastructure (CSDI, 1995) | Taiwan National Land Surveying and Mapping Information Web Map Service | (Bucci et al., 2022) | Geology |
| Fault lines | | | European Geological Data Infrastructure (ISPRA-SGI, 2021) | Distance to faults |
| NDVI | Chinese Academy of Science Discipline Data Centre for Ecosystem (30m) (Yang et al., 2019) | | Calculated using Landsat 8 | NDVI |
| LULC | (Potapov et al., 2022) | | | LULC |
| Rainfall | Hong Kong Observatory GEO | CMORPH Climate Data Record (CDR) | | Rainfall |

| | | | |
|---|---|---|---|
| | Data Portal | | |
| Road network | OpenStreetMap (OSM) | | Distance to roads |
| River network | | | Distance to rivers |

## 2.3 AlphaEarth Embeddings

In addition to LCFs, Google AE embeddings, which are updated on annual basis, were used in this study as an alternative representation of environmental conditions for landslide susceptibility modelling. The AE datasets were accessed and processed using the Google Earth Engine (GEE) platform. For Nantou County and Emilia-Romagna, the AE embeddings were temporally matched to the corresponding landslide inventories (Table1), with embeddings from 2022 and 2023 used, respectively. For Hong Kong, the retained landslide inventory spans 2000-2019 and is cumulative, the 2019 AE embeddings were therefore used to represent the environmental conditions corresponding to the most recent year covered by the inventory. The spatial coverage of the AE embeddings was masked to match that of the LCF-based datasets for each study area. All AE embeddings were resampled to a spatial resolution of 30 m × 30 m to ensure consistency with the LCF inputs.

# 3 Methods

## 3.1 Data Pre-processing

### 3.1.1 Principal Component Analysis for AlphaEarth Embeddings

Since AE embeddings provide a high-dimensional representation of environmental conditions derived from multi-source remote sensing data, the original 64-dimensional feature space (AE all 64 bands) may contain correlated information, as multiple embedding dimensions can respond to similar underlying environmental patterns. Rather than assuming such correlations to be inherently detrimental, this characteristic reflects the rich and redundant nature of learned representations. To provide a linearly constrained alternative representation of AE embeddings for comparative analysis, principal component analysis (PCA) was applied as a supplementary transformation to the original AE features (AE PCA bands), allowing a direct comparison between full-dimensional and reduced-dimensional AE inputs in subsequent LSM (Table 3).

PCA is a widely used unsupervised technique that projects the original feature space onto a set of orthogonal components (Salih Hasan & Abdulazeez, 2021), providing a compact representation that preserves the dominant variance structure of the original embeddings. Given a standardised AE embedding matrix $X \in \mathbb{R}^{n \times p}$, where $n$ denotes the number of samples and $p = 64$ represents the number of original AE embedding dimensions, PCA seeks a linear transformation that projects $X$ into a lower-dimensional space. This is achieved by eigenvalue decomposition of the covariance matrix $C$, defined as:

$$C = \frac{1}{n-1} X^T X \tag{1}$$

where the eigenvectors of $C$ define the principal component directions and the corresponding eigenvalues indicate the amount of variance explained by each component. The transformed data matrix $Z$ is obtained as:

$$Z = XW \tag{2}$$

where $W \in \mathbb{R}^{p \times k}$ is the projection matrix composed of the top $k$ eigenvectors corresponding to the largest eigenvalues.

### 3.1.2 Multicollinearity Diagnostics

Multicollinearity among input variables may affect the robustness and stability of the model. To examine the degree of linear dependence among selected predictors, multicollinearity diagnostics were conducted using tolerance and variance inflation factor (*VIF*). For a given predictor $x_i$, tolerance $(T)$ is defined as:

$$T = 1 - R_i^2 \tag{3}$$

where $R_i^2$ is the coefficient of determination obtained by regressing $x_i$ on all remaining predictors. The corresponding VIF is given by:

$$VIF = \frac{1}{T} \tag{4}$$

In this study, multicollinearity diagnostics were applied to the LCFs group and the AE representations derived through PCA. These two groups of variables constitute predictors under a linear framework, for which variance-based diagnostics such as tolerance and variance inflation factor are appropriate. By contrast, multicollinearity diagnostics were not applied to the original 64-dimensional AE embeddings. The AE embeddings represent a learned high-dimensional feature space that does not naturally align with the assumptions underlying variance-based linear diagnostics. Instead, the performance of the original AE embeddings was evaluated directly through modelling results (Table 3).

**Table 3**. Conceptual comparison of different input representations used in this study

| Category | LCFs | AE (PCA bands) | AE (all 64 bands) |
| --- | --- | --- | --- |
| Origin | Physically or empirically defined variables | Derived from AE embeddings via PCA | Learned from multi-source remote sensing data |
| Representation | Explicit environmental factors | Linear projection of learned representations | Learned high-dimensional representations |
| Role in this study | Conventional baseline representation | Linearly constrained alternative for comparison | Primary representation for data-driven modelling |
| Dimensionality | Low to moderate (explicitly defined) | Reduced-dimensional (selected principal components) | High-dimensional (all 64 embedding features) |
| Evaluation strategy | Statistical diagnostics and model performance | Statistical diagnostics and model performance | Model performance only |

## 3.2 Modelling Approaches

The overall workflow of this study is illustrated in Fig. 2. Three representative deep learning architectures were adopted, including a one-dimensional convolutional neural network (CNN1D), a two-dimensional convolutional neural network (CNN2D), and a Vision Transformer (ViT). These models were selected to provide complementary modelling approaches that differ in how they represent and process the input information. Specifically, CNN1D represents a feature-based model that operates only on the attributes of the target pixel, CNN2D extends this representation by incorporating local spatial context through two-dimensional convolution, and ViT provides an attention-based alternative for modelling relationships among pixels within the same spatial neighbourhood. Rather than providing an exhaustive comparison of deep learning architectures, these models were used to examine whether the relative performance of LCFs and AE remained consistent across these different modelling strategies.

The CNN1D architecture was designed to operate on vectorised feature representations, in which each sample is described by a one-dimensional feature sequence. In this formulation, only the attributes associated with the landslide or non-landslide pixel were considered, without explicitly incorporating the surrounding spatial context. Accordingly, the input to the CNN1D model takes the form $X \in \mathbb{R}^{1\times p}$, where $p$ denotes the number of input features. Specifically, $p = 14$ for the LCFs-based input, $p$ corresponds to the number of retained principal components for the AE (PCA bands), and $p = 64$ for the AE (all 64 bands). Within this framework, one-dimensional convolutional filters were applied along the feature dimension to extract high-level representative patterns from the raw input feature vectors, providing a feature-based baseline without modelling spatial context.

The CNN2D architecture extended the feature-based representation by incorporating local spatial context surrounding each sample. This approach treated a landslide or non-landslide location not as an isolated pixel, but as the centre of a localised geomorphological unit. Each sample was constructed as a multi-channel tensor $X \in \mathbb{R}^{h\times w\times p}$, formed by extracting an image patch centred on the landslide or non-landslide points. In this study, the spatial window size was fixed as $11\times 11$ for all study areas, with channel $p$ varied according to the data source (LCFs or AE embeddings). By applying two-dimensional convolutional kernels, the CNN2D model learned spatial and feature representations within local neighbourhoods relevant to landslides.

In contrast to the fixed receptive fields of convolution-based architectures, the Vision Transformer (ViT) adopts a self-attention paradigm to process the local spatial windows. In this study, the same local spatial windows used for CNN2D were represented as a sequence of flattened pixel-level tokens. To preserve the spatial arrangement destroyed by flattening, learnable positional embeddings were added to the token sequence. The resulting token sequence was then processed by transformer encoder layers, allowing relationships among all pixel-level tokens to be modelled through self-attention. Under this formulation, ViT provided a mechanism to capture spatial relationships within local windows without relying on convolutional locality assumptions.

Detailed implementations of the three model architectures are illustrated in Fig. 3. Hyperparameters were optimised using the exhaustive grid search strategy, with the candidate settings summarised in Table 5. Batch size and initial learning rate were selected for tuning because of their direct influence on optimisation stability and convergence. Adam was used for optimisation with cross-entropy loss, and the learning rate was adjusted using a polynomial decay schedule with a decay power of 0.95. The optimal configuration for each experiment was selected according to the validation AUC. This procedure ensured that the three architectures and different input representations were evaluated under a consistent optimisation framework. All models were formulated as binary classifiers to distinguish landslide samples (label = 1) from non-landslide samples (label = 0). The balanced sample dataset for each study area was randomly divided into training and validation subsets at a ratio of 70:30. The split was performed at the sample level without additional spatial or temporal separation. No independent test set was used in the primary experiments, and model performance was evaluated using the validation dataset.

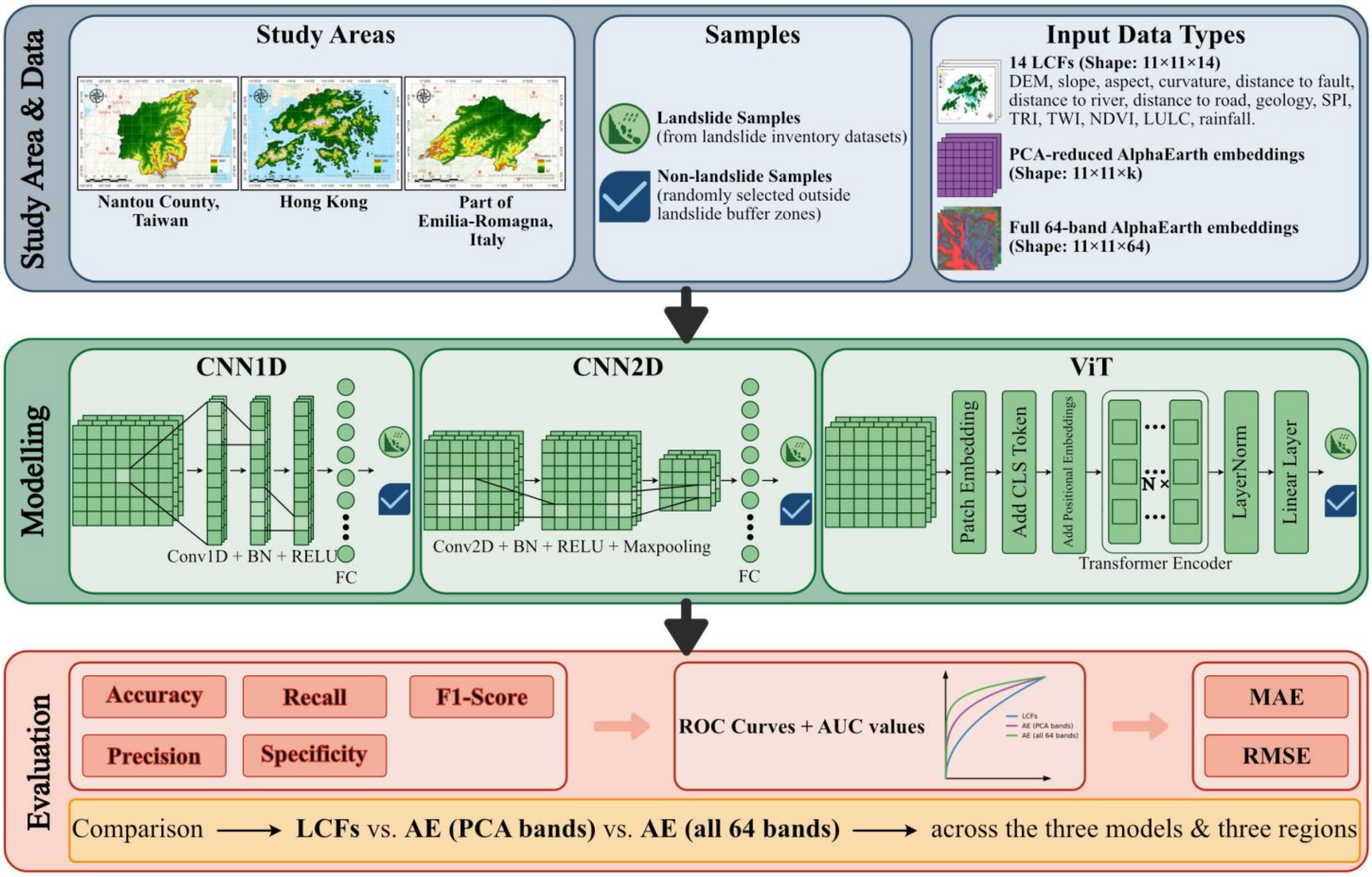

**Fig. 2**. Overall workflow of the study.

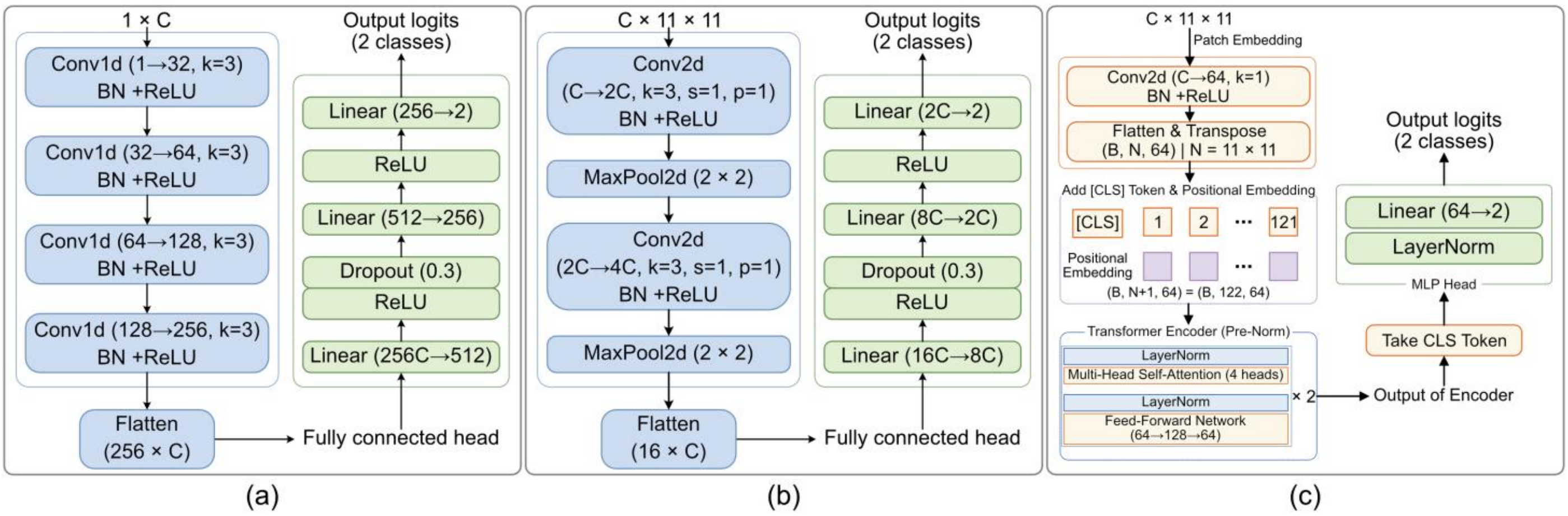

**Fig. 3**. Detailed architectures of the three deep learning models adopted in this study: (a) CNN1D, (b) CNN2D, and (c) ViT.

**Table 5**. Hyperparameter settings, training configuration, and computational environment used in this study.

| Settings | Values |
|---|---|
| Batch sizes | 512, 1024, 2048 |
| Initial learning rates | $10^{-5}$, $10^{-4}$, $10^{-3}$, $10^{-2}$, $10^{-1}$ |
| Learning rate scheduler | Polynomial decay (PolyLR), with decay power $p = 0.95$ |
| Optimizer | Adam |
| Loss function | Cross-entropy loss |
| Number of epochs | 200 |
| Early stopping | Monitored on validation loss, with tolerance $1 \times 10^{-4}$ and patience of 10 epochs. |
| CPU | AMD EPYC 7282 16-Core Processor |
| GPU | NVIDIA GeForce RTX 4090 D (24 GB VRAM) |
| Memory | 128 GB |
| Hard disk | 2 TB |

## 3.3 Evaluation Strategy

To comprehensively assess the performance of landslide susceptibility models, multiple evaluation strategies were adopted in this study. All evaluation metrics were computed based on the validation dataset. First, model performance was assessed using confusion-matrix-based metrics, including accuracy, precision, recall (also known as sensitivity), F1-score, and specificity (Fig. 4). These metrics provide complementary perspectives on class-wise prediction reliability and the balance between correctly and incorrectly classified landslide and non-landslide samples. In addition, receiver operating characteristic (ROC) curves were used to evaluate the discriminatory ability of the models across different decision thresholds, which is a widely adopted approach in landslide susceptibility mapping (Alqadhi et al., 2024; Moghimi et al., 2024; Abdelkader & Csámer, 2025). The area under the ROC curve (AUC) was calculated as a threshold-independent indicator of model performance, reflecting the overall capability of each model to distinguish between landslide and non-landslide samples.

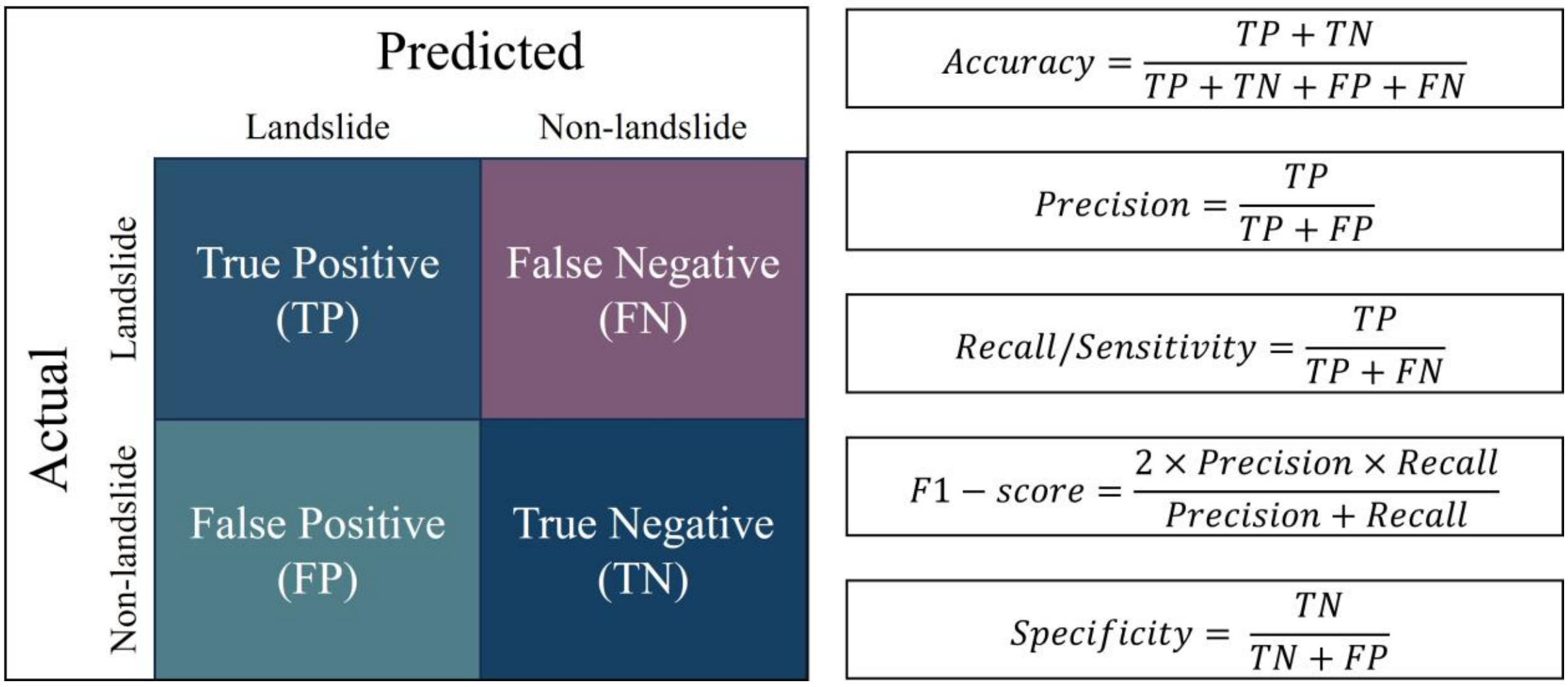

**Fig. 4**. Confusion matrix and formulas of the evaluation metrics used in this study.

In addition, binary cross-entropy (log loss) was calculated to evaluate the quality of the probabilistic predictions. Log loss assigns larger penalties to incorrect predictions made with high confidence, providing a classification-specific measure of probabilistic prediction performance (Formula 5).

$$Log\ Loss = -\frac{1}{N}\sum_{i=1}^{N}[y_i \ln(p_i) + (1 - y_i)\ln(1 - p_i)] \tag{5}$$

where $N$ denotes the total number of samples in the validation dataset, $y_i$ represents the observed label of the $i-th$ sample, with landslide and non-landslide encoded as 1 and 0, respectively, and $p_i$ denotes the predicted probability of the landslide class for the corresponding sample.

As complementary measures of prediction error, regression-style error metrics were further employed to evaluate the continuous susceptibility outputs. Specifically, mean absolute error (MAE) and root mean square error (RMSE) were calculated to quantify the deviation between predicted susceptibility values and observed landslide labels (Formulae 6 and 7). MAE measures the average magnitude of prediction errors, whereas RMSE assigns greater weight to larger errors.

$$RMSE = \sqrt{\frac{\sum_{i=1}^{M}(y_i - y_i')^2}{M}} \tag{6}$$

$$MAE = \frac{\sum_{i=1}^{M}|y_i - y_i'|}{M} \tag{7}$$

where $M$ denotes the total number of samples in the validation dataset, $y_i$ represents the label of the $i$-th sample (with landslide and non-landslide encoded as binary values), and $y_i'$ denotes the corresponding continuous susceptibility value of the $i$-th sample.

### 3.4 SHAP-based Feature Attribution

To examine how individual input features contributed to model predictions, SHapley Additive exPlanations (SHAP) was employed as a post-hoc interpretation method (Shapley, 1953; Lundberg & Lee, 2017). SHAP attributes a model prediction to individual input features based on the Shapley value from cooperative game theory. Rather than assessing a feature in isolation, its contribution is evaluated across different combinations of the remaining features, thereby accounting for its marginal influence under different feature contexts. For a given sample $x$, the SHAP value of feature $i$ can be expressed as:

$$\phi_i(x) = \sum_{S \subseteq F \setminus \{i\}} \frac{|S|!\,(M-|S|-1)!}{M!} [v_x(S \cup \{i\}) - v_x(S)] \tag{8}$$

where $F$ denotes the complete set of $M$ input features, $S$ is a subset of $F$ excluding feature $i$, and $v_x(S)$ represents the model output obtained when the features in $S$ are considered. The resulting SHAP values satisfy the additive relationship:

$$f(x) = E[f(X)] + \sum_{i=1}^{M} \phi_i(x) \tag{9}$$

where $E[f(X)]$ is the expected model output and $\phi_i(x)$ represents the contribution of feature $i$ to the prediction for sample $x$. Positive and negative SHAP values indicate contributions that respectively increase or decrease the predicted landslide susceptibility relative to the expected model output.

To characterise feature importance at the dataset level, the absolute SHAP values were aggregated across the analysed samples. The global importance of feature $i$ was therefore calculated as:

$$I_i = \frac{1}{N} \sum_{j=1}^{N} |\phi_i(x_j)| \tag{10}$$

where $N$ is the number of samples included in the interpretation analysis. Features with larger mean absolute SHAP values were considered to have a greater overall influence on model predictions.

## 4 Results

### 4.1 Principal Components for AlphaEarth Embeddings

Fig. 5 shows the cumulative explained variance of the AE embeddings as a function of the number of retained principal components for the three study areas. Across all regions, the cumulative explained variance increased sharply with the first few components and then gradually levelled off. Based on this pattern, a cumulative explained variance threshold of 90% was adopted to determine the number of components retained for each study area, and the selected components were used as a reduced-dimensional representation of the AE embeddings in subsequent modelling experiments. Under this criterion, 13 principal components were sufficient to reach 90% cumulative variance in Nantou County (Fig. 5a) and Hong Kong (Fig. 5b). In Emilia, however, a slightly larger number of components was required, with about 15 principal components needed to reach the same cumulative variance level (Fig. 5c).

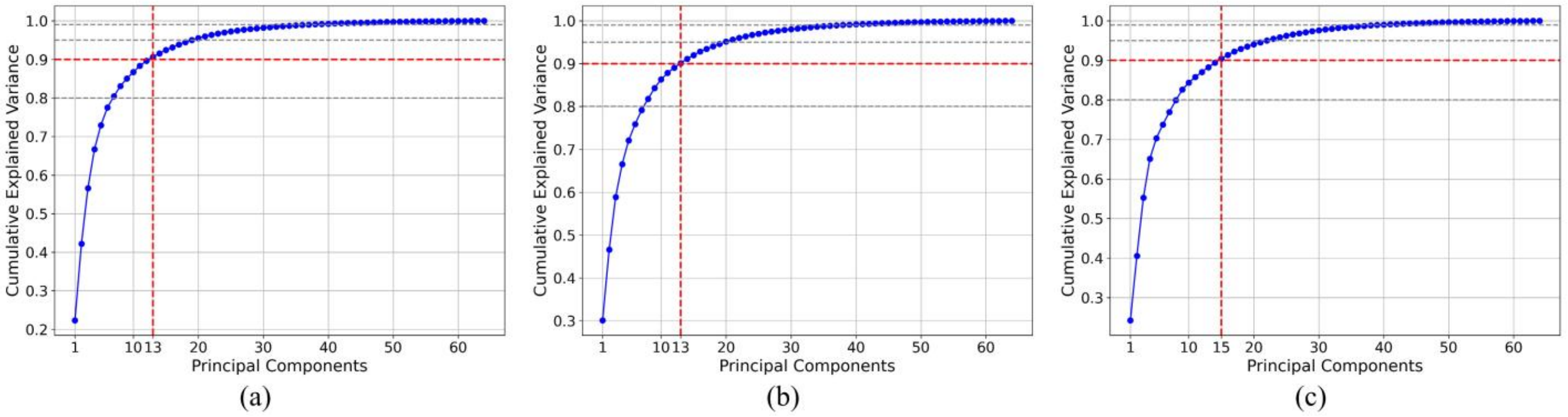

**Fig. 5**. Cumulative explained variance of principal components derived from AlphaEarth embeddings in **(a)** Nantou; **(b)** Hong Kong; **(c)** Emilia.

## 4.2 Multicollinearity Diagnostics Results

Tolerance and VIF were calculated for both the LCFs and the AE (PCA bands) across the three study areas (Fig. 6). In general, a tolerance value greater than 0.1 and a VIF value lower than 10 were considered indicative of acceptable levels of multicollinearity. Fig. 6(a) presents the tolerance and VIF values for the 14 LCFs. All tolerance values exceeded the threshold of 0.1, with the highest value observed for aspect in Hong Kong (0.996) and the lowest for rainfall in Emilia (0.267). Correspondingly, VIF values for all LCFs remained well below 10 across the three regions, ranging from 1.004 to 3.743. These results indicate that none of the LCFs exhibit severe multicollinearity, suggesting that they can be jointly used for model training without introducing critical redundancy. Fig. 6(b) shows the multicollinearity diagnostics for the AE (PCA bands). For all retained principal components, both tolerance and VIF values satisfied the adopted thresholds in the three study areas. This indicates that the PCA-based dimensionality reduction effectively alleviated linear dependencies among features while preserving the dominant variance structure of the original AE embeddings.

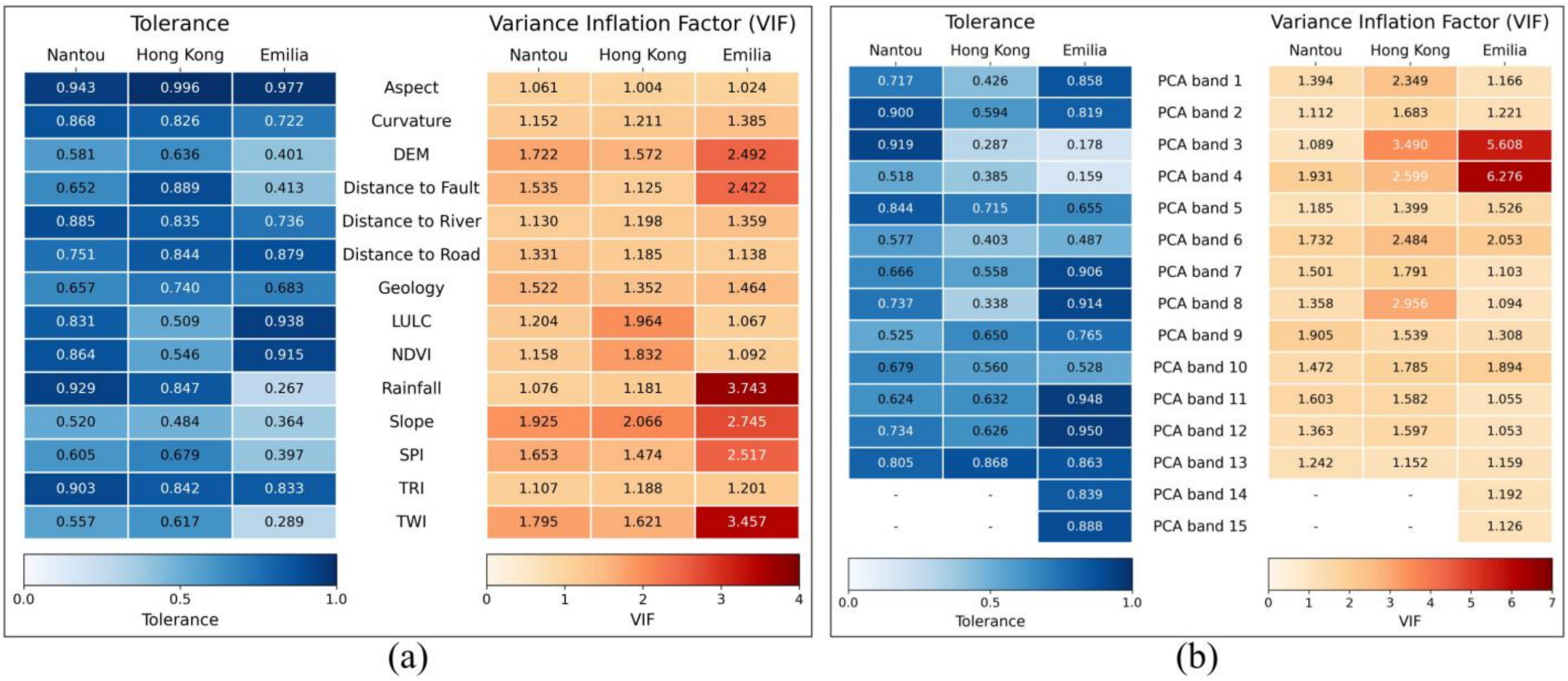

**Fig. 6**. Multicollinearity diagnostics using Tolerance and VIF: **(a)** LCFs; **(b)** PCA-based AlphaEarth components.

## 4.3 Comparison of Landslide Susceptibility Maps

To illustrate the different spatial characteristics of landslide susceptibility maps derived from AE-based predictors and conventional LCFs, Nantou County is presented here as a representative example. Fig. 7 presents landslide susceptibility maps for Nantou County generated using the three models (CNN1D, CNN2D, and ViT) and three input datasets (LCFs, AE with PCA bands, and AE with all 64 bands). All susceptibility maps were classified into five levels, namely "very low", "low", "moderate", "high", and "very high", using the natural breaks method, and are displayed together with the landslide inventory to facilitate spatial comparison.

At the regional scale, the overall spatial pattern of landslide susceptibility was broadly consistent across all experiments. Higher susceptibility was predominantly concentrated in the eastern part of Nantou County, gradually decreasing westward toward areas classified mainly as "very low" susceptibility. This pattern is consistent with the general topographic setting of Nantou, where rugged mountainous terrain is primarily distributed in the east, while gentler landscapes and built-up areas dominate the western part of the county.

Despite these similar macroscopic patterns, noticeable differences can be observed in the detailed spatial distribution of landslide susceptibility when different data types were used. Across all modelling approaches, susceptibility maps generated from LCFs-based inputs tend to show relatively gradual transitions between classes, with a considerable proportion of areas were classified into "low", "moderate", or "high" susceptibility levels. While LCF-based models were able to delineate most high-risk zones associated with landslide points in mountainous terrain, their ability to distinguish landslides occurring in generally low-risk environments was relatively limited (e.g., the zoomed area in the western part of Nantou County in Fig. 7), as such landslides were not always clearly highlighted by elevated susceptibility levels like "high" or "very high".

In contrast, susceptibility maps derived from AE features, including both AE (PCA bands) and AE (all 64 bands), exhibited a more sensitive response at known landslide locations, regardless of the specific model employed. Most landslide points were directly associated with "very high" susceptibility values, with fewer intermediate classifications such as "moderate" or "high". Notably, landslides located within low-risk surroundings were more clearly emphasised in the AE-based maps. Visually, the susceptibility patterns obtained using AE (PCA bands) and AE (all 64bands) were highly similar, showing nearly identical macroscopic distributions as well as fine-scale spatial textures, suggesting that the dimensionality reduction can preserve the key information required for LSM.

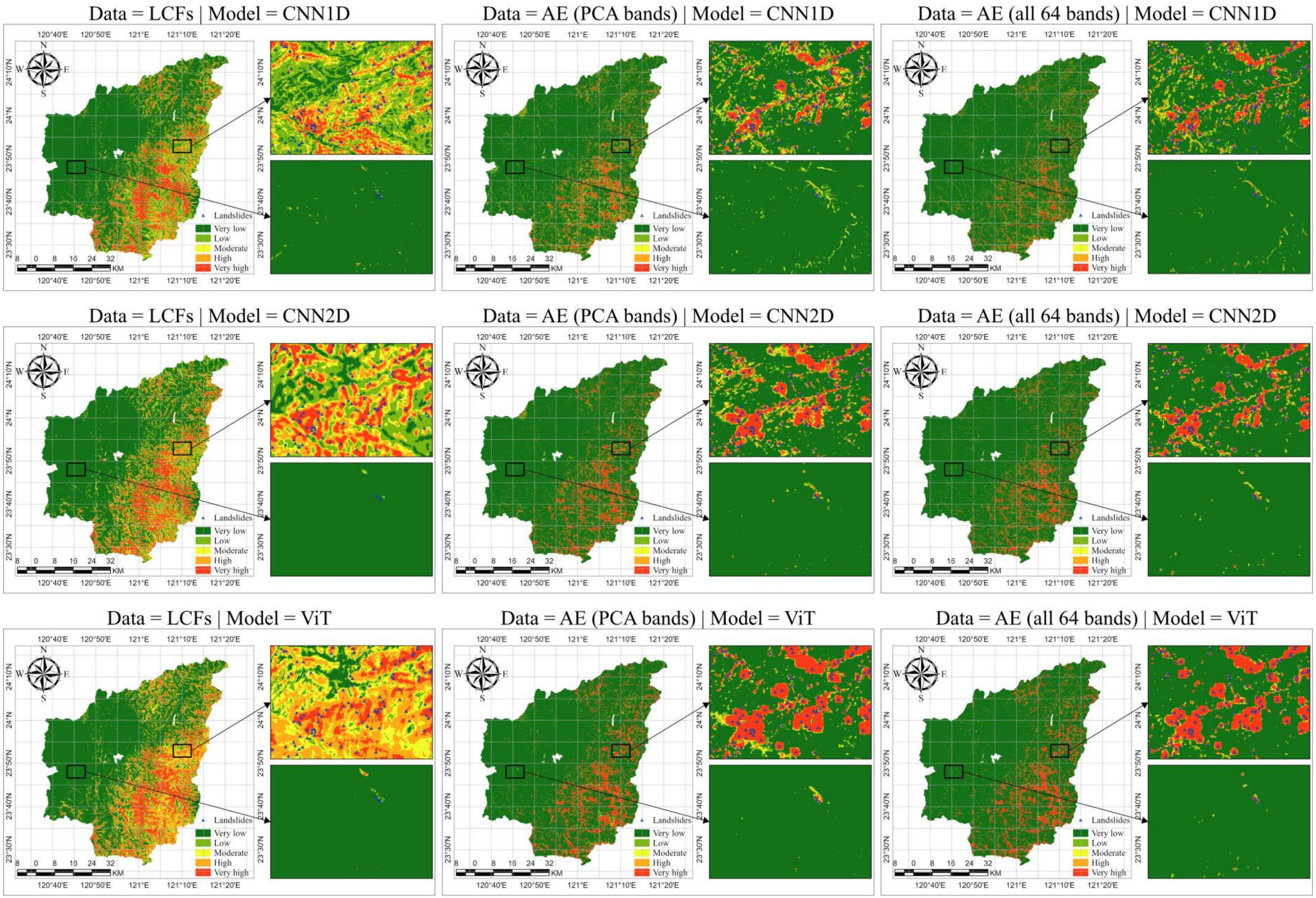

**Fig. 7**. Comparison of landslide susceptibility maps derived from LCFs and AE using different models in Nantou, Taiwan. **LCF**: landslide conditioning factors. **AE**: AlphaEarth Embeddings.

## 4.4 Quantitative Comparison of LCFs and AlphaEarth Embeddings

To further compare the predictive performance of LSM using LCFs and AE-derived predictors, quantitative evaluation metrics were calculated. Fig. 8 summarises the performance of the three models across the three study areas using five evaluation metrics, including accuracy, precision, recall, specificity, F1-score, and log loss. Across all study areas and model architectures, models trained with AE-derived predictors consistently achieved higher and more stable metric values than those trained with LCFs. The same performance ordering (LCFs < AE PCA bands < AE all 64 bands) was observed across the three regions and for all three models, suggesting that the gains reflect a general advantage of AE-derived representations rather than an isolated effect tied to one model or one study area.

Taking Nantou as a representative example, models trained with LCFs typically yielded performance scores of around 80% across the five metrics, with moderate variations among CNN1D, CNN2D, and ViT. In contrast, when AE-based predictors were used, both AE (PCA bands) and AE (all 64 bands) led to a pronounced increase in performance, with most metrics exceeding 95% for all three models. Within the AE-based results, the use of AE (all 64 bands) generally resulted in slightly higher scores across the evaluated metrics, indicating that the complete AE set may provide a more informative representation for landslide prediction.

A similar performance pattern was observed in Hong Kong and Emilia, although the magnitude of improvement varied among regions. Specifically, the performance gains associated with AE features were most pronounced in Nantou, followed by Emilia, while the improvement in Hong Kong was comparatively minor. This regional contrast was particularly evident for CNN1D, where the use of AE (PCA bands) led to only marginal improvement over LCFs in Hong Kong, whereas a more noticeable gain was achieved when AE (all 64 bands) were employed. In contrast, for Nantou and Emilia, even AE (PCA bands) already provided a substantial performance increase relative to LCFs.

In addition, ROC curves and AUC values provide a complementary and threshold-independent evaluation of model discrimination performance. Fig. 9 presents the ROC curves and corresponding AUC values for the three models across the three study areas using LCFs, AE (PCA bands), and AE (all 64 bands). Across all regions and models, AE-based predictors consistently yielded ROC curves that lay closer to the upper-left corner than those obtained using LCFs, reflecting stronger separation between landslide and non-landslide samples. In Nantou, AUC values increased from approximately 0.90 for LCF-based models to around 0.99 when AE features were used, with only minor differences between AE (PCA bands) and AE (all 64 bands). Similar improvements were observed in Hong Kong and Emilia, where AE-based inputs consistently achieved higher AUC values across all models. In terms of model comparison, CNN2D and ViT generally exhibited slightly higher AUC values than CNN1D for the same input data.

Fig. 10 presents the error distributions and the corresponding MAE and RMSE values for Nantou County under different model and dataset combinations. Across all three models, predictions based on LCFs exhibited relatively wide error dispersion, with errors spread over a broad range and larger deviations from zero. Consistently, the associated MAE and RMSE values remained comparatively high, with MAE values generally around 0.25 and RMSE values exceeding 0.35, reflecting less stable prediction behaviour. In contrast, the use of AE-derived predictors led to a pronounced reduction in both error magnitude and dispersion. For CNN1D, MAE decreased to approximately 0.05 when AE (PCA bands) were used and further to around 0.04 with AE (all 64 bands), accompanied by a substantial reduction in RMSE. Similar patterns were observed for CNN2D and ViT. Comparable error reductions were also evident in Hong Kong and Emilia (Appendix Fig. A1 and Fig. A2), where AE-based inputs consistently produced tighter error clusters centred close to zero.

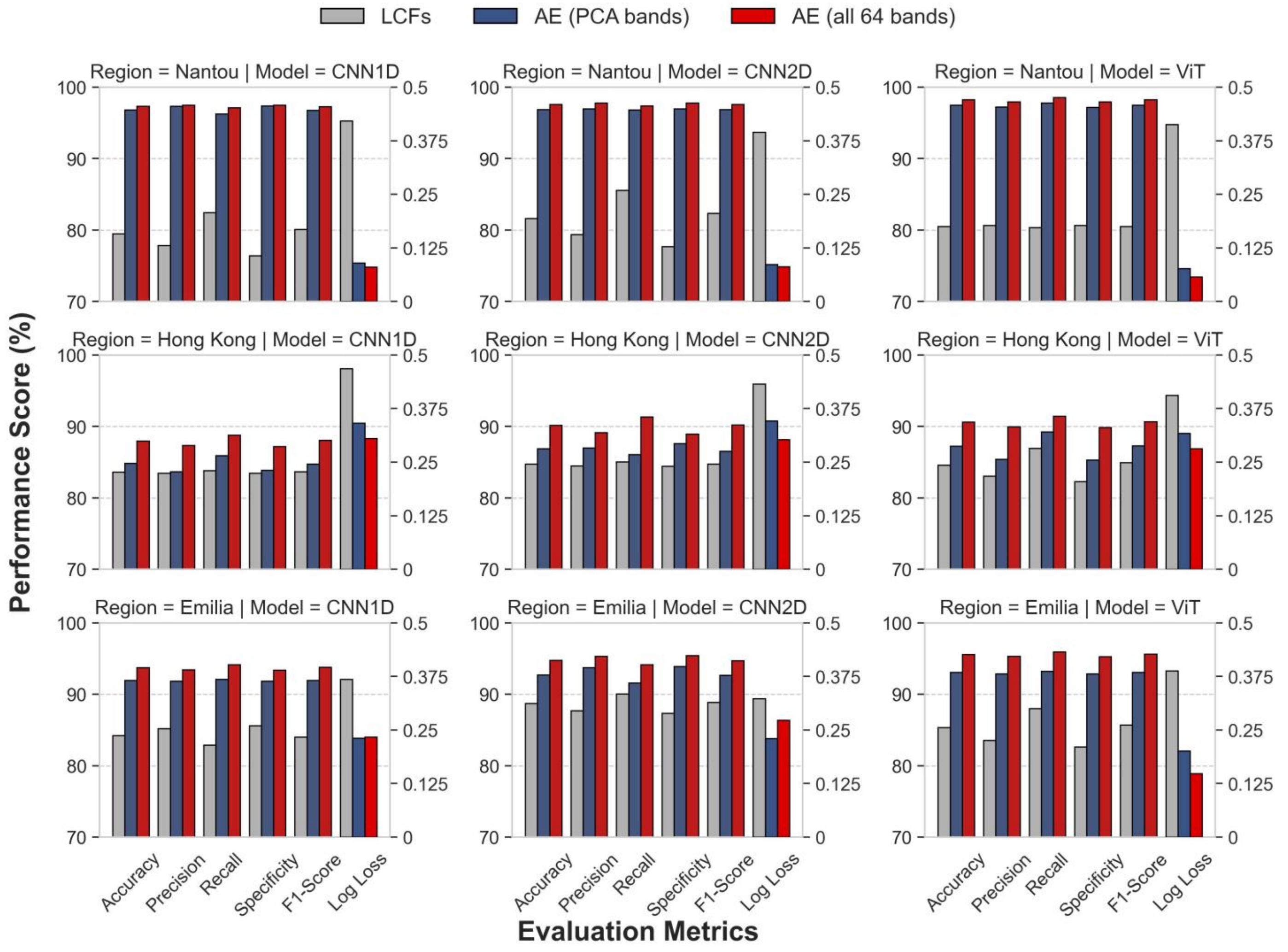

**Fig. 8**. Performance comparison of different data sources across three deep learning models and study areas. **LCF**: landslide conditioning factors. **AE**: AlphaEarth Embeddings.

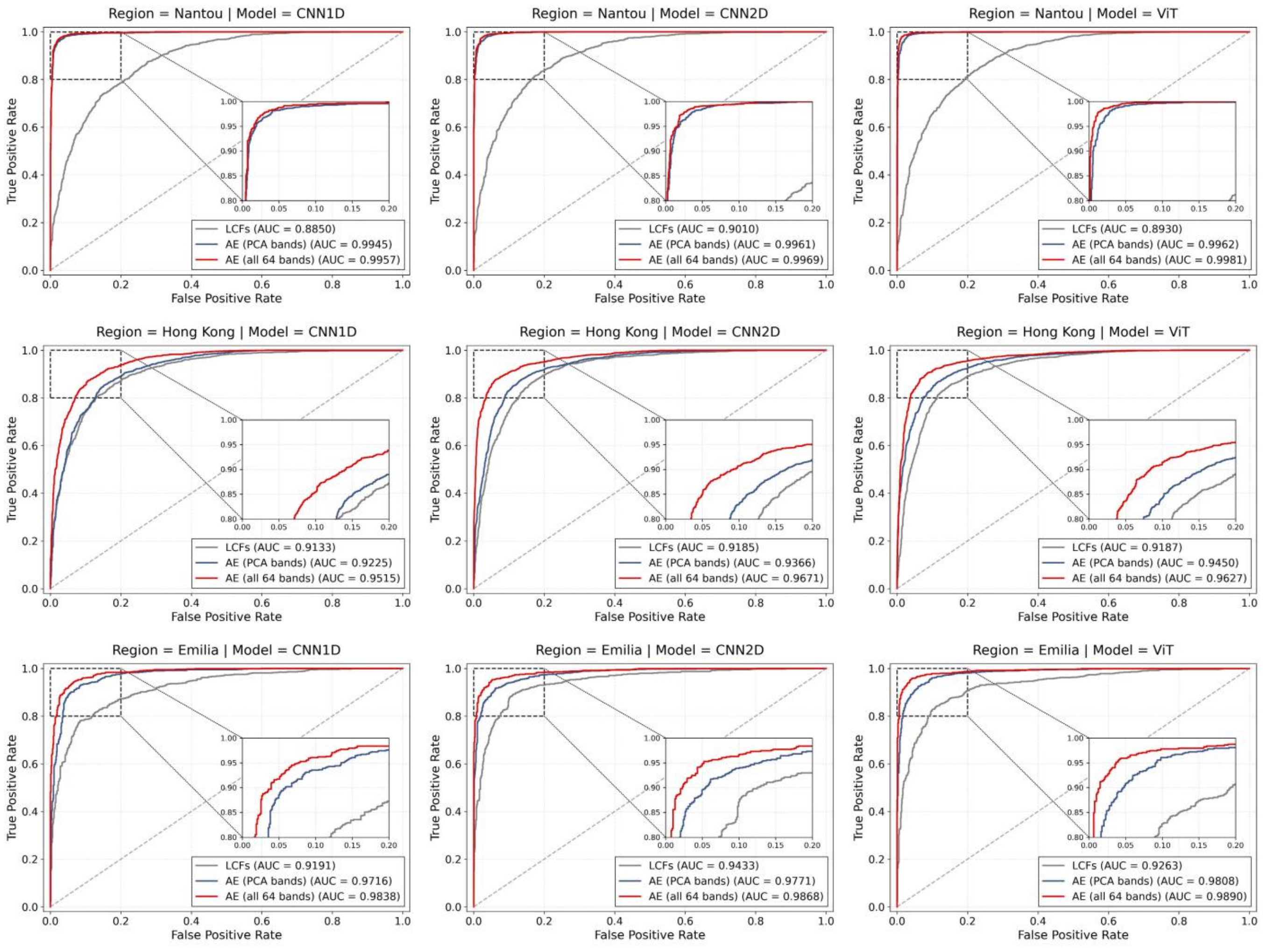

**Fig. 9**. Receiver Operating Characteristic (ROC) curves and Area under the Curve (AUC) comparison of different data sources across three deep learning models and study areas. **LCF**: landslide conditioning factors. **AE**: AlphaEarth Embeddings.

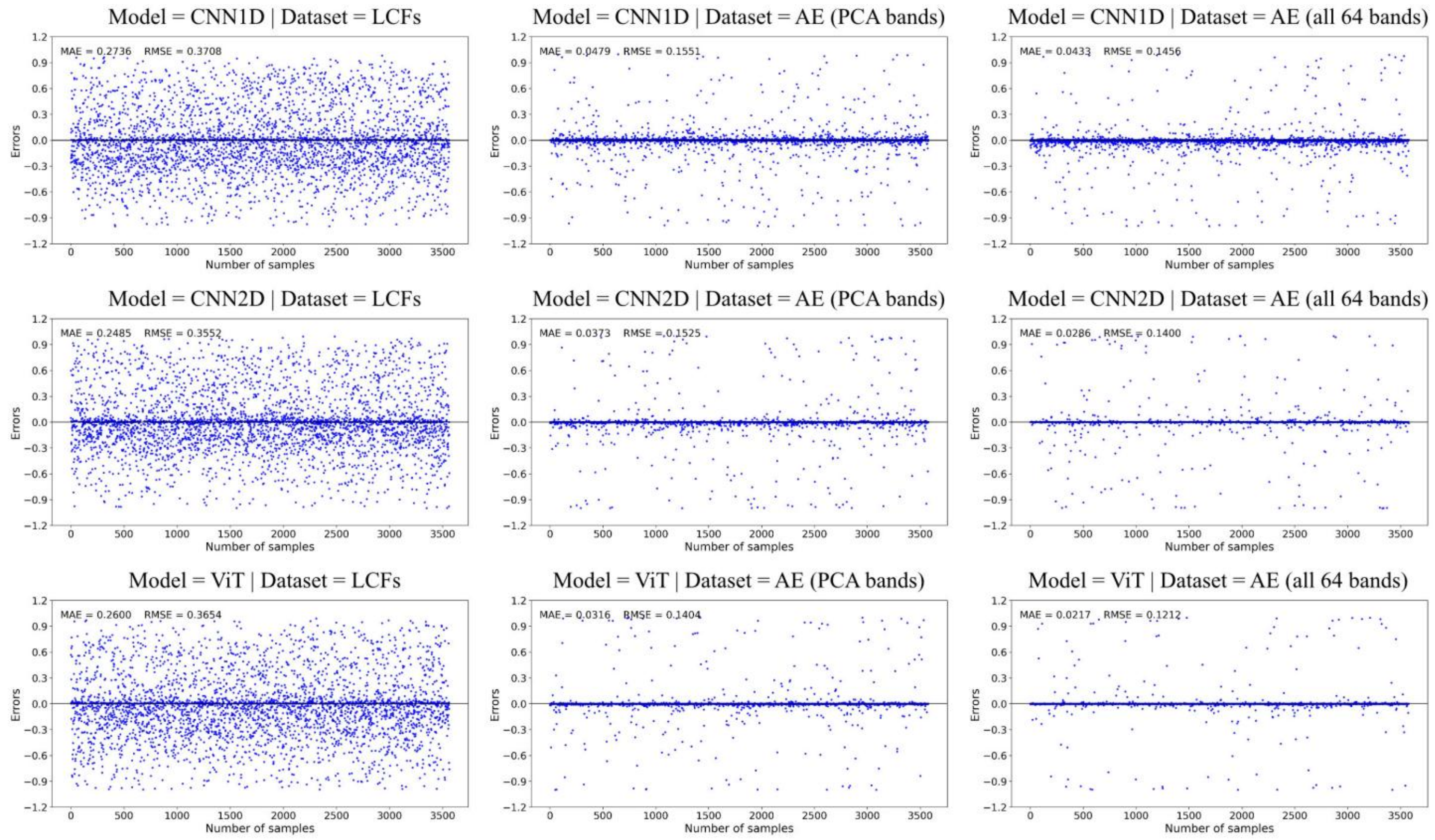

**Fig. 10**. Error distribution of different data sources across three deep learning models in Nantou, Taiwan. **LCF**: landslide

conditioning factors. **AE**: AlphaEarth Embeddings.

## 4.5 SHAP-based attribution of AlphaEarth embeddings

To further investigate which bands of the AE embeddings contributed most strongly to landslide susceptibility, SHAP analysis was conducted for the ViT models trained using AE (all 64 bands). Fig. 11 presents the 15 highest-ranked bands in each study area according to their mean absolute SHAP values. The SHAP summary plots additionally show the direction and distribution of individual feature contributions.

The SHAP results reveal both cross-regional consistency and marked spatial heterogeneity in the contribution of individual AE bands. Bands 15, 17, and 48 were consistently ranked among the top 15 in all three study areas, with band 48 appearing within the top five in every region. This repeated importance suggests that several AE dimensions may encode latent environmental information that is broadly relevant to landslide susceptibility across different geographical settings. In contrast, most of the other highly ranked bands differed considerably among Nantou, Hong Kong, and Emilia-Romagna, indicating that the information used by the models is not spatially uniform but partly reflects local environmental and landslide conditions. It should be emphasised that AE bands are learned latent embedding dimensions rather than directly observable environmental variables. Their SHAP importance therefore identifies influential components of the representation but does not establish the physical meaning or geomorphological mechanism represented by each band. Consequently, the common importance of bands such as 48 should be interpreted as evidence of a potentially transferable latent signal.

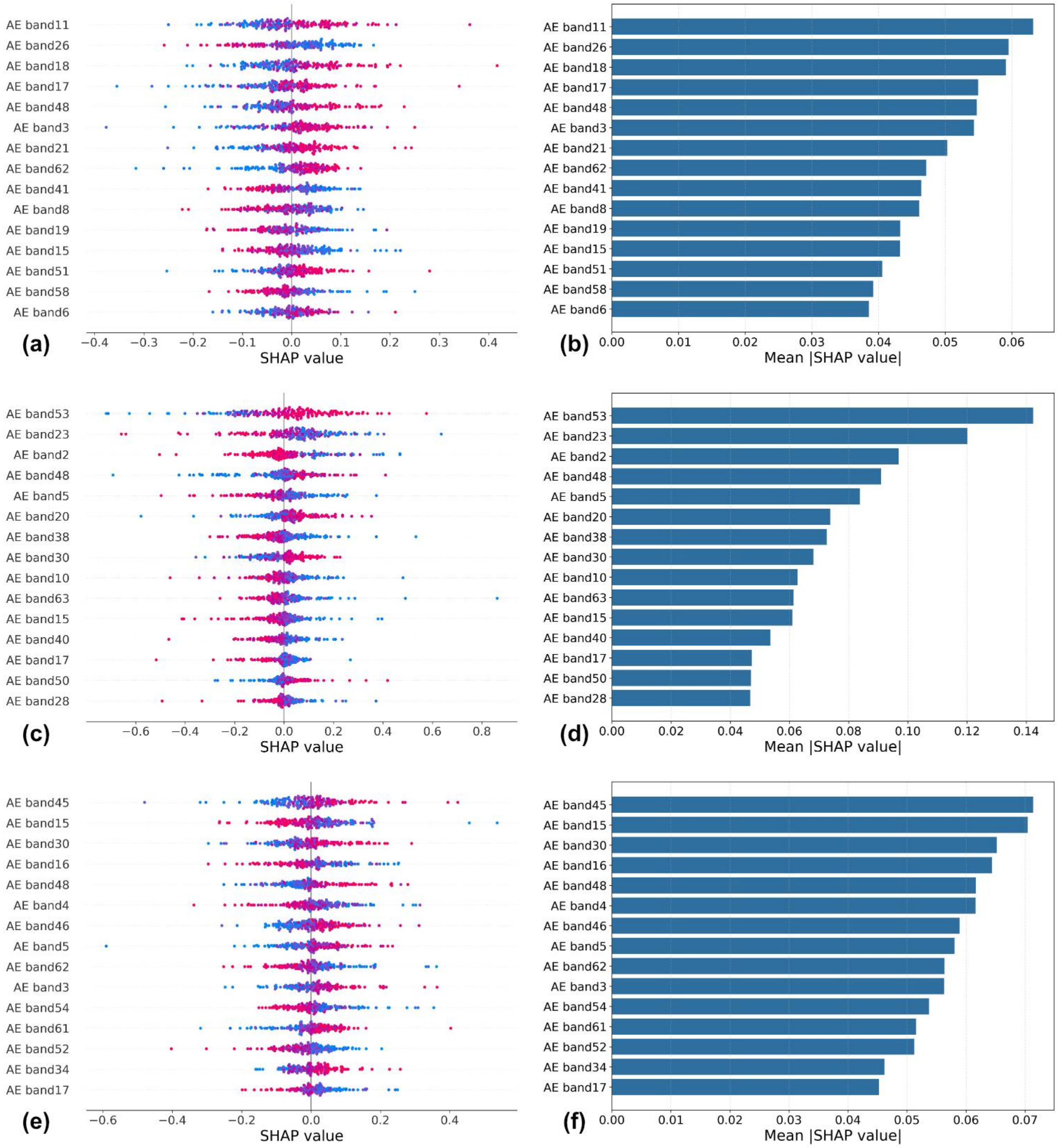

**Fig. 11**. SHAP-based attribution of the top 15 AlphaEarth embedding bands for the ViT models using beeswarm and mean absolute SHAP feature-importance plots: (a) and (b): Nantou; (c) and (d): Hong Kong; (e) and (f): Emilia.

# 5 Discussion

## 5.1 Evaluation of Predicted Susceptibility Classes at Recorded Landslide Locations

To further evaluate the spatial agreement between the predicted susceptibility patterns and observed landslide occurrences, the landslide inventory points were overlaid with the categorised susceptibility maps. From an LSM perspective, stronger spatial agreement between predicted high-risk zones and observed landslide occurrences is desirable. Therefore, a larger share of landslide points falling within the "very high" susceptibility class is typically regarded as a more reasonable and convincing outcome.

To construct this comparison, the susceptibility class of the 30 m grid cell containing each landslide point was extracted, and the numbers of landslide points assigned to the five susceptibility levels were subsequently counted and expressed as percentages of the total number of landslides. Han and Semnani (2025) adopted a similar category-based comparison and further accounted for potential positional uncertainty by assigning the highest susceptibility class within a predefined neighbourhood around each reported landslide location. In the present study, landslide locations were directly matched to the corresponding 30 m grid cells. Although this provides a straightforward one-to-one correspondence between the landslide locations and predicted susceptibility classes, positional uncertainty in the original inventories was not explicitly modelled. Therefore, some uncertainty may remain in the assigned susceptibility classes, particularly for landslide points located close to class boundaries. Fig. 12 summarises the distribution of existing landslide points across five susceptibility levels for the three study areas and three models, using LCFs, AE (PCA bands), and AE (all 64 bands).

A clear contrast was observed between LCF-based and AE-based inputs. With LCFs, landslide points were spread more broadly across classes: while many points still fell within "high" or "moderate” susceptibility zones, a noticeable portion remains in "low" and, in some cases, "very low" classes. This more dispersed allocation is consistent with the map-based observations in Section 4.3, where LCF-based outputs tended to show wider intermediate-susceptibility belts and sometimes provided less distinct highlighting of some individual landslide locations. In contrast, when AE-derived predictors were used, landslide points were much more concentrated in the "very high" class across all three models and all three study areas, with a relatively smaller fraction assigned to lower levels. This concentration was strongest for AE (all 64 bands), whereas AE (PCA bands) showed a similar but slightly less pronounced shift toward the top class, consistent with a slight information loss introduced by dimensionality reduction.

This pattern, combined with metric- and map-based evidence reported in Section 4, suggests that AE-based LSM outputs are generally more decisive and reliable in identifying areas associated with landslides and are more suitable for practical screening and mapping. A plausible explanation is that AE embeddings provide an integrated description of the landscape derived from large-scale training, which may capture subtle environmental patterns that are difficult to express through a limited set of manually selected LCFs. By contrast, LCF-based predictors are often constructed as separate thematic layers or proxy variables (e.g., generalised lithology classes and distance-to measures). While effective for outlining broad susceptibility patterns, these representations may be less sensitive to localised conditions at specific landslide sites.

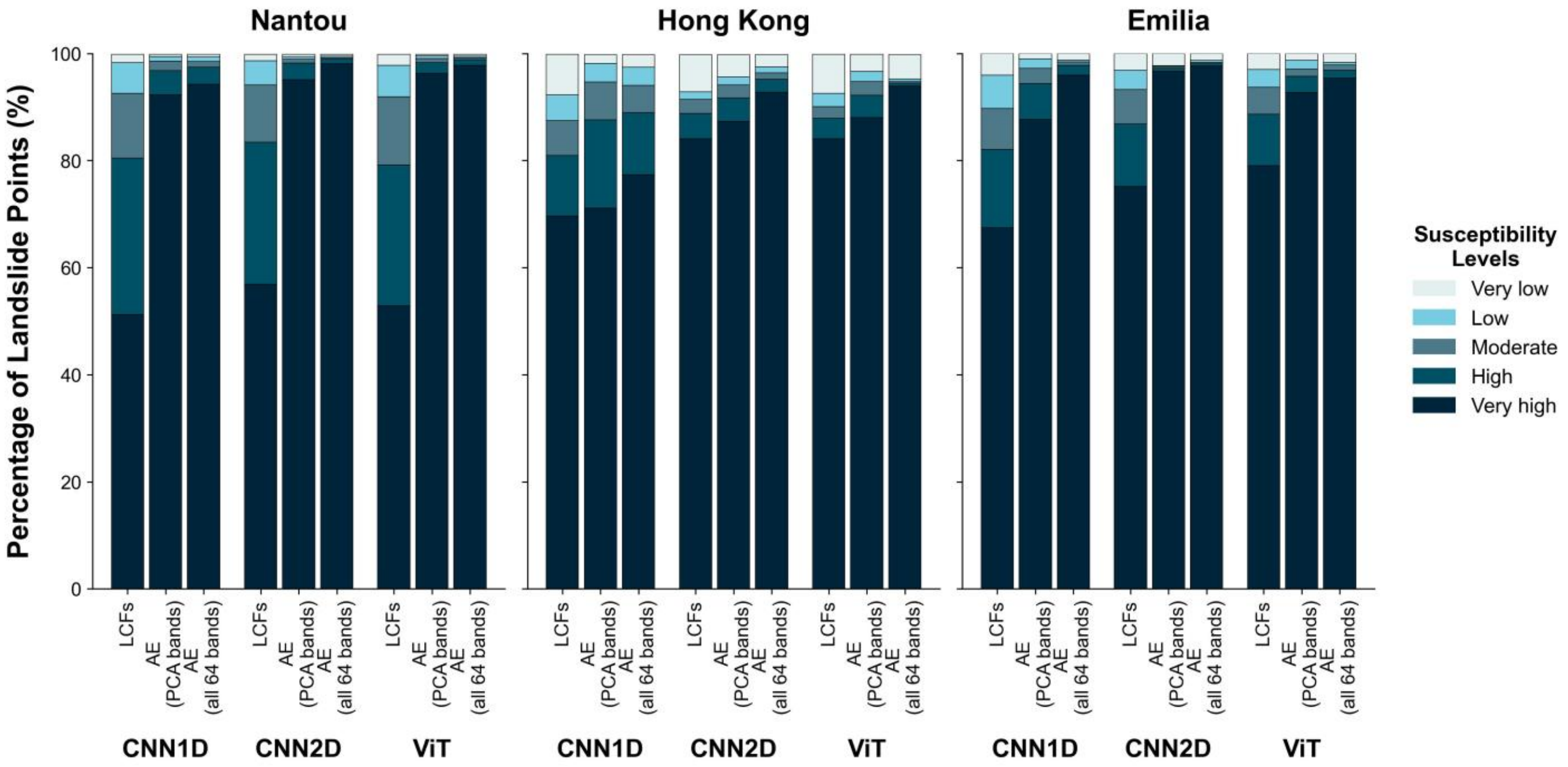

**Fig. 12**. Percentage of landslide points falling into each susceptibility level for different models and input datasets across three study areas. **LCF**: landslide conditioning factors. **AE**: AlphaEarth Embeddings.

## 5.2 Robustness of AE-based susceptibility modelling

To further examine the robustness of the observed performance differences, an additional spatial five-fold evaluation was conducted. As shown in Fig. 13, the results were highly consistent with those obtained from the original evaluation. Across all three study areas and model architectures, the mean AUC consistently followed the same ordering of "LCFs < AE (PCA bands) < AE (all 64 bands)". The advantage of AE-derived predictors was particularly pronounced in Nantou, while clear improvements were also maintained in Hong Kong and Emilia-Romagna.

The fold-wise results also showed relatively limited variability for the AE-based models. In most model-region combinations, both AE representations produced smaller standard deviations than the corresponding LCF-based models, indicating more stable performance across different data partitions. Notably, the full 64-band representation retained the highest mean AUC for all nine model-region combinations. Together with the original results, these findings provide further evidence that the performance gains associated with AE embeddings are reproducible across different spatial data partitions and are not restricted to a particular model architecture or study area.

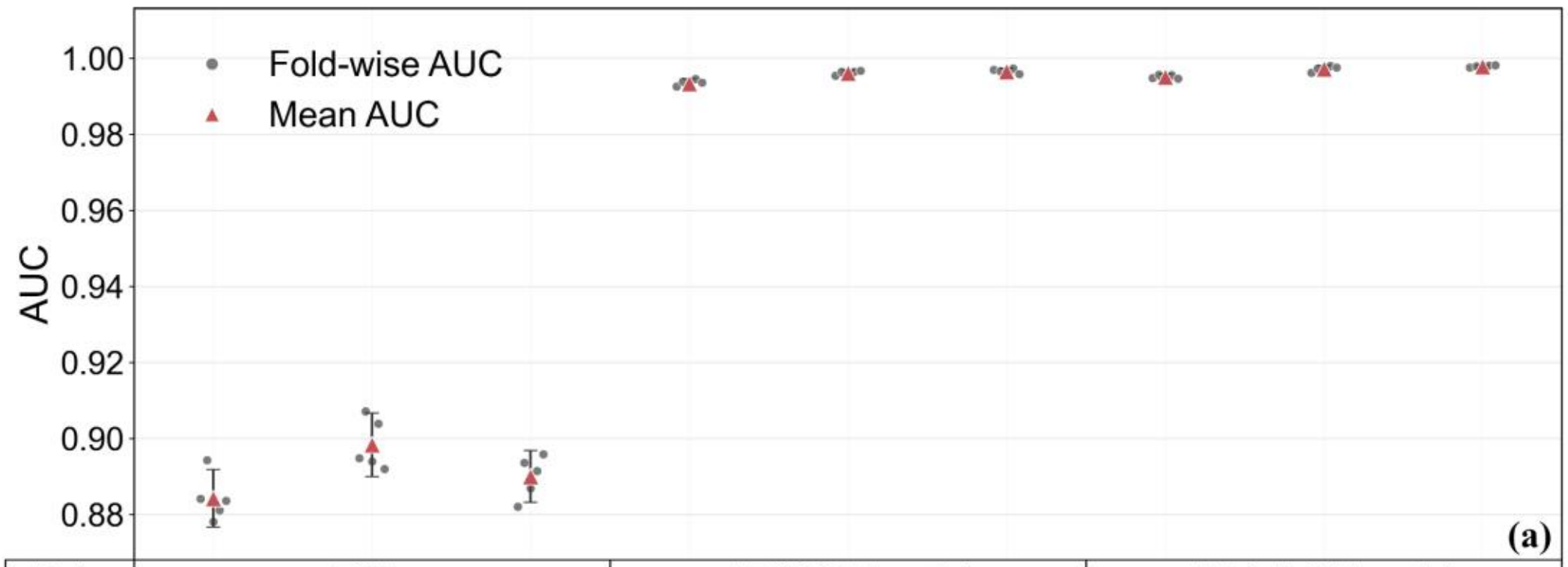

| Data | LCFs | | | AE (PCA bands) | | | AE (all 64 bands) | | |
|---|---|---|---|---|---|---|---|---|---|
| Model | CNN1D | CNN2D | ViT | CNN1D | CNN2D | ViT | CNN1D | CNN2D | ViT |
| Mean | 0.8842 | 0.8983 | 0.8900 | 0.9933 | 0.9961 | 0.9965 | 0.9951 | 0.9972 | 0.9978 |
| SD | 0.0061 | 0.0067 | 0.0055 | 0.0010 | 0.0006 | 0.0007 | 0.0005 | 0.0007 | 0.0003 |

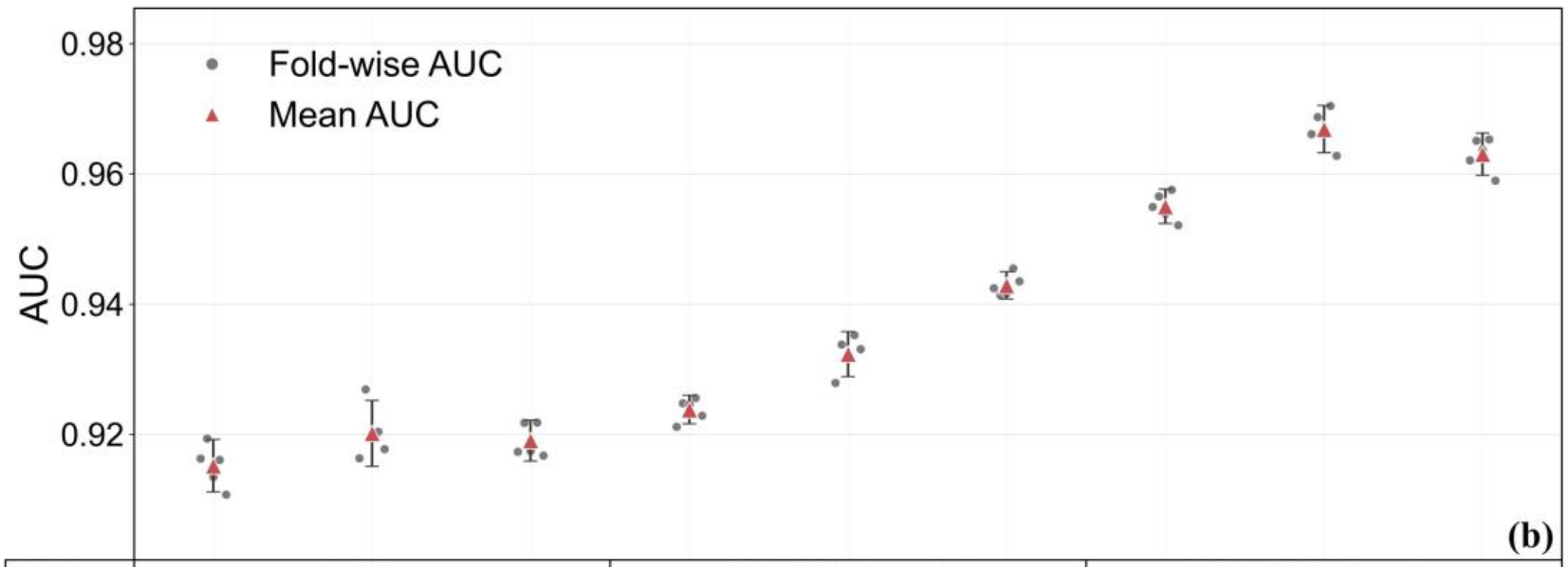

| Data | LCFs | | | AE (PCA bands) | | | AE (all 64 bands) | | |
|---|---|---|---|---|---|---|---|---|---|
| Model | CNN1D | CNN2D | ViT | CNN1D | CNN2D | ViT | CNN1D | CNN2D | ViT |
| Mean | 0.9152 | 0.9201 | 0.9190 | 0.9238 | 0.9323 | 0.9429 | 0.9550 | 0.9668 | 0.9630 |
| SD | 0.0033 | 0.0041 | 0.0025 | 0.0018 | 0.0028 | 0.0017 | 0.0021 | 0.0029 | 0.0026 |

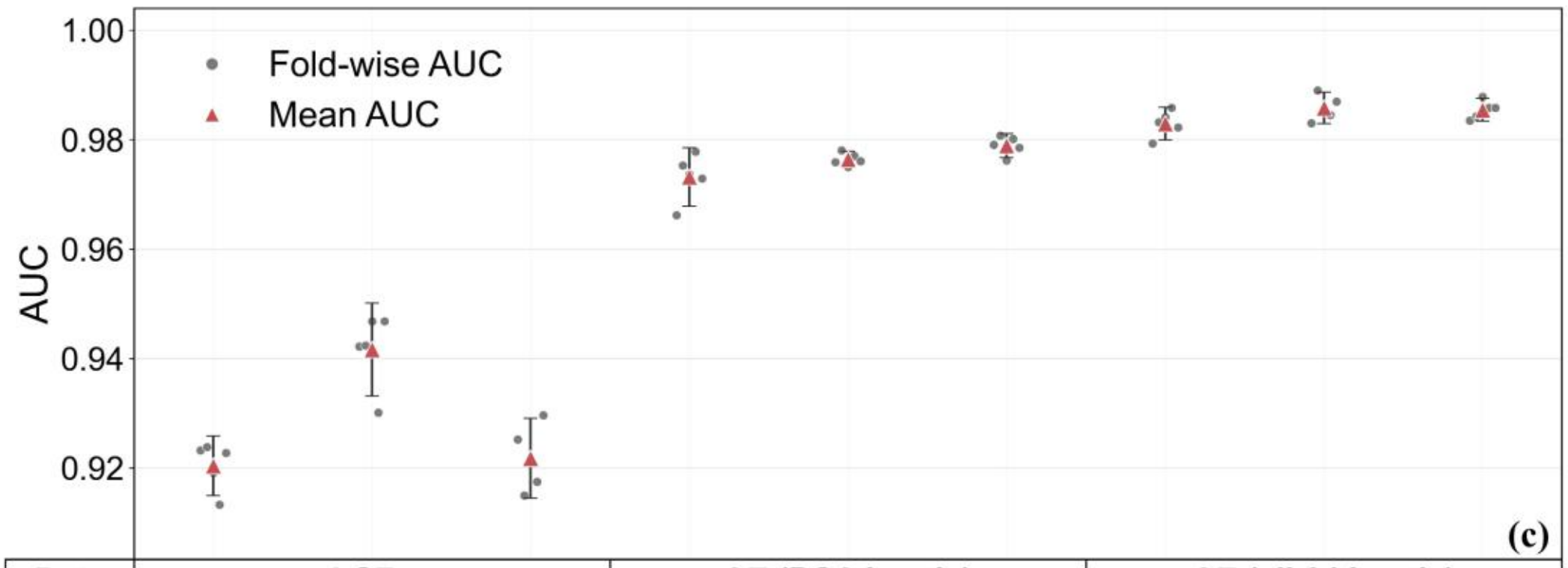

| Data | LCFs | | | AE (PCA bands) | | | AE (all 64 bands) | | |
|---|---|---|---|---|---|---|---|---|---|
| Model | CNN1D | CNN2D | ViT | CNN1D | CNN2D | ViT | CNN1D | CNN2D | ViT |
| Mean | 0.9205 | 0.9417 | 0.9218 | 0.9732 | 0.9764 | 0.9789 | 0.9829 | 0.9858 | 0.9855 |
| SD | 0.0044 | 0.0068 | 0.0059 | 0.0043 | 0.0012 | 0.0018 | 0.0024 | 0.0023 | 0.0017 |

**Fig. 13**. Five-fold AUC results based on spatial partitioning for models using LCFs, AE (PCA bands), and AE (all 64 bands) in (a) Nantou, (b) Hong Kong, and (c) Emilia.

## 5.3 Performance Gains of AlphaEarth Embeddings Across Study Areas

Although AE-derived predictors consistently improved LSM performance across all three study areas, the magnitude of improvement was not spatially uniform. In particular, the gains observed in Nantou and Emilia were noticeably larger than those in Hong Kong. One possible factor associated with this difference is the temporal correspondence between the AE embeddings and the landslide inventories, although its influence cannot be isolated from the current experiments. AE features are generated annually and thus reflect the surface conditions of a specific year. In Nantou and Emilia, the landslide inventories are either event-based or updated at an annual or near-annual frequency. In Taiwan, for example, landslides are systematically interpreted and annually updated by the Taiwan Ministry of Agriculture through visual interpretation of high-resolution imagery. As a result, the recorded landslide locations are closely aligned with the contemporaneous land-surface conditions captured by the AE embeddings. This temporal consistency likely enhances the ability of AE features to characterise landslide-prone environments, resulting in more pronounced performance gains.

In contrast, the retained ENTLI records for Hong Kong span 2000-2019. Consequently, the surface conditions represented by the AE data may not fully match those earlier landslide records, particularly where vegetation recovery, land-cover change, or human intervention has occurred. Such temporal mismatch may introduce additional uncertainty in the relationship between landslide locations and AE-derived surface representations. Previous studies have similarly shown that uncertainty in landslide inventories can affect LSM performance (Huang, Li, et al., 2024; Huang, Mao, et al., 2024). However, the present study did not include a controlled experiment specifically designed to quantify this effect, and the smaller performance improvement in Hong Kong cannot be attributed to temporal mismatch alone. Regional differences in environmental conditions, inventory characteristics, and the information captured by AE embeddings may also contribute to the observed variation. Future studies using landslide inventories with more precise temporal information could directly compare temporally matched and mismatched AE inputs to determine the effect of temporal consistency.

## 5.4 Limitations and Future Research Directions

This study demonstrates the potential of Google AE embeddings for LSM across multiple regions and model architectures. However, the comparison between AE embeddings and conventional LCFs should be interpreted with some caution. AE is generated by a pretrained foundation model that integrates information from multiple Earth observation sources, whereas the conventional baseline in this study consists of 14 manually selected conditioning factors. Therefore, the observed performance gains may partly reflect the richer information contained in the AE representation rather than the representation method alone. The results should be viewed as evidence of the effectiveness of AE as an alternative input representation for LSM, rather than proof of its inherent superiority over conventional LCFs. Future studies could provide more controlled comparisons by incorporating a broader range of LCFs.

Another limitation is that AE was not compared with other recent Earth observation foundation models or self-supervised remote sensing representations. The proliferation of this field has produced a growing range of pretrained representations, such as Prithvi (Jakubik et al., 2023) and TESSERA (Feng et al., 2026). These approaches differ in their pretraining data, sensor modalities, spatial and temporal representation, and learning objectives, and may therefore capture different information relevant to landslide susceptibility. Future studies may benchmark AE against other alternatives to determine whether the observed improvements are specific to AE or represent a broader advantage of foundation model based representations for LSM.

In addition, the current AE dataset spans a limited temporal range (2017-2025). As additional years become available, AE embeddings may enable more systematic investigation of interannual variability in landslide susceptibility, particularly in years affected by major triggering events such as earthquakes or extreme rainfall. Longer AE time series may also support the analysis of the temporal evolution of landslide susceptibility, including gradual changes, cumulative effects, and post-event recovery processes. A key challenge in this context is the availability of landslide inventories with sufficient temporal precision to align with the AE time frame. Without such temporal matching, aggregating landslide samples across multiple years may introduce uncertainties, potentially reducing the effectiveness of temporally explicit surface representations.

Furthermore, AE embeddings are derived from a globally trained foundation model and thus provide a broadly consistent description of land-surface characteristics across regions. This property suggests potential advantages for LSM in settings where landslide samples are sparse or unevenly distributed. Future research could explore this potential through cross-region transfer experiments, evaluating whether AE-based models can retain predictive performance under limited local training data.

# 6 Conclusion

This study examined the adoption of Google AE embeddings as an alternative feature representation for data-driven LSM. By comparing AE-based predictors with conventionally-employed LCFs across three study areas, namely Nantou County, Hong Kong, and Emilia, using three deep learning models, including CNN1D, CNN2D, and ViT, AE-based inputs consistently achieved more reliable and stronger predictive performance than LCF-based approaches.

Across all regions and models, both types of AE representation, including the AE (PCA bands) and the AE (all 64 bands), outperformed LCF-based inputs across all evaluation metrics, including accuracy, precision, recall, specificity, F1-score, and AUC. Models using the complete set of 64 AE bands generally showed further performance improvements, suggesting that retaining richer embedding information can be beneficial for LSM. In addition, AE-based models exhibited more stable error distributions, characterised by lower MAE and RMSE values, as well as a clearer spatial correspondence between predicted "very high" susceptibility zones and actual recorded landslide occurrences. Notably, in areas where landslides were embedded within surroundings characterised by generally low susceptibility levels, AE-based results were able to identify localised high-susceptibility zones, indicating enhanced sensitivity to landslide-prone conditions.

However, the magnitude of these performance gains varies among the study areas, with larger gains observed in Nantou and Emilia than in Hong Kong. Temporal differences between AE embeddings and landslide inventories may partly contribute to this variation, although their influence cannot be quantitatively isolated from other regional differences using the current experiments. SHAP analysis further revealed substantial regional heterogeneity in the contribution of individual AE bands. Nevertheless, bands 15, 17, and 48 were consistently ranked among the 15 most influential bands across all three study areas, with band 48 appearing within the top five in every region.

Despite these encouraging findings, several aspects warrant further investigation. Future research could examine whether compact subsets of informative AE bands can retain comparable predictive performance. More controlled comparisons with broader sets of conventional LCFs and other foundation model embeddings are also needed to better understand the relative value of AE for LSM. In addition, the potential of AE embeddings for LSM in data-scarce regions, as well as their application to analysing interannual and long-term susceptibility dynamics as longer AE time series become available, remains to be explored.

### CRediT authorship contribution statement

**Y.C.**: Conceptualisation, methodology, software, formal analysis, data curation, investigation, visualisation, writing - original draft, writing—review and editing. **Q.Z.**: Conceptualisation, methodology, formal analysis, data curation, software, investigation, writing—review and editing. **L.F.**: Conceptualisation, methodology, software, formal analysis, data curation, investigation, supervision, writing—review and editing.

### Acknowledgements

This research was supported by the Xi'an Jiaotong-Liverpool University Postgraduate Research Scholarship under Grant FOSA2312049.

### Declaration of competing interest

The authors declare that they have no known competing financial interests or personal relationships that could have appeared to influence the work reported in this paper.

### Data availability

The datasets used and/or analysed during the current study will be made available from the corresponding author on reasonable request.

# Appendix A

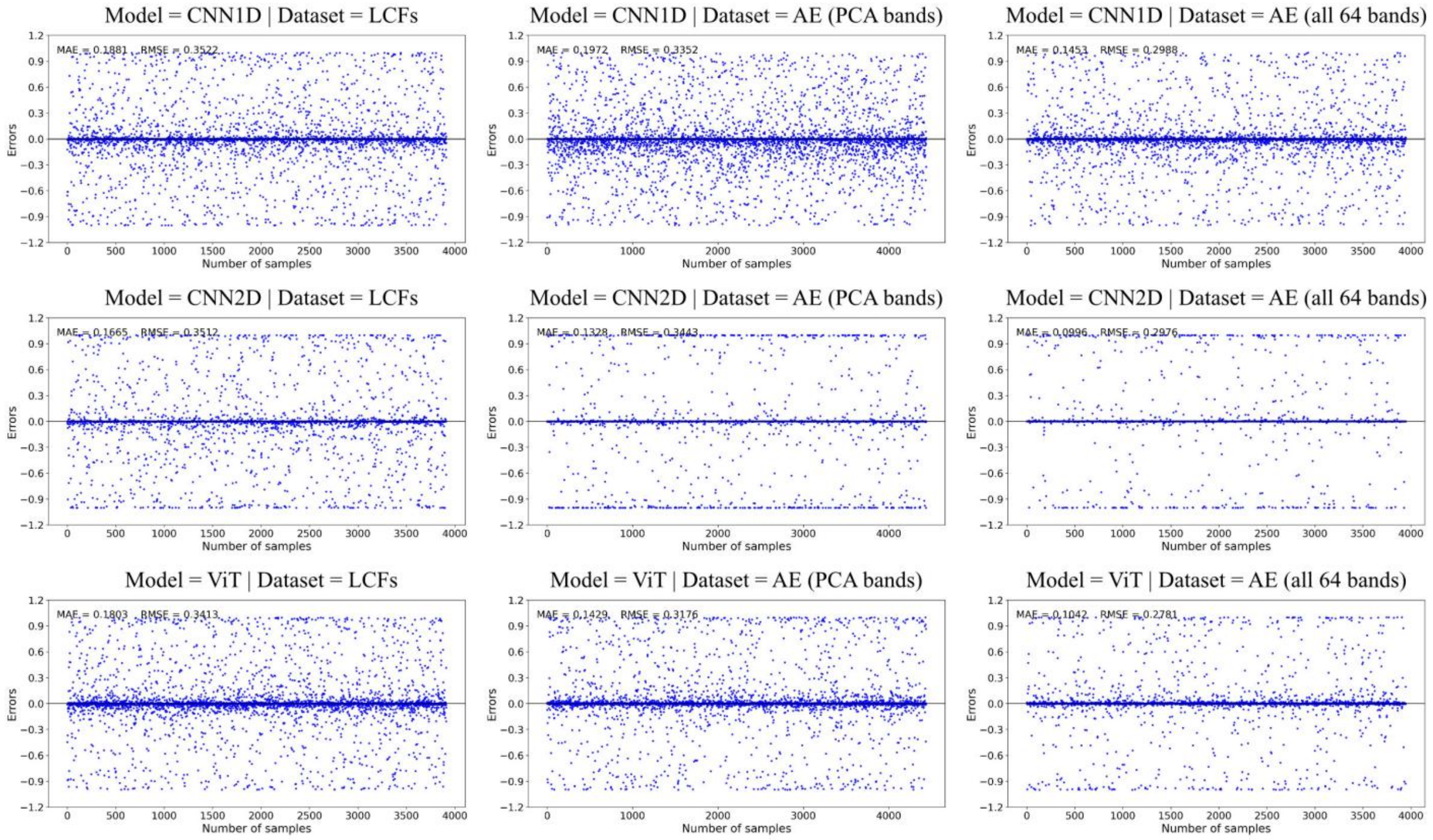

**Fig. A1**. Error distribution of different data sources across three deep learning models in Hong Kong. **LCF**: landslide conditioning factors. **AE**: AlphaEarth Embeddings.

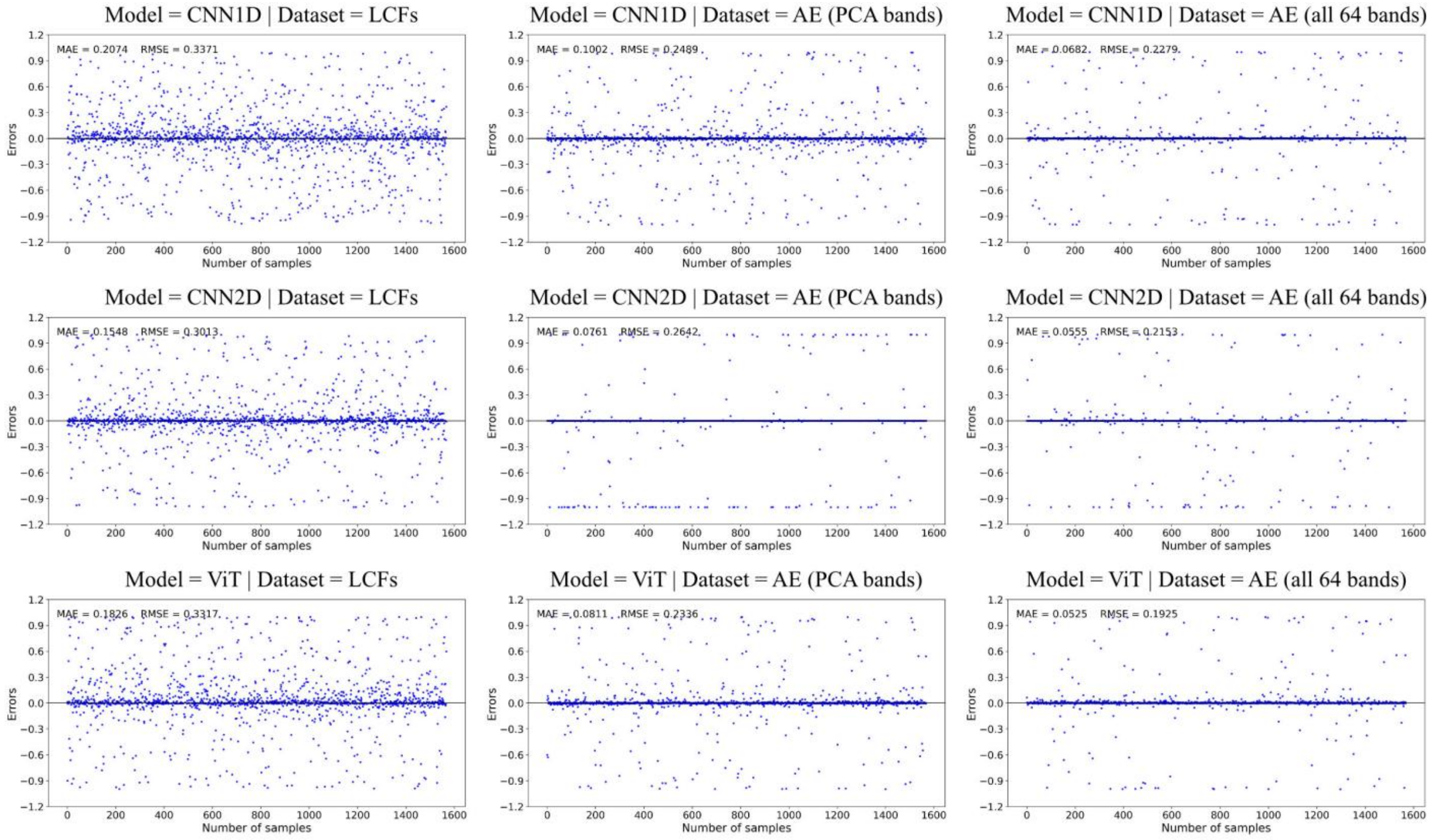

**Fig. A2**. Error distribution of different data sources across three deep learning models in Emilia, Italy. **LCF**: landslide conditioning factors. **AE**: AlphaEarth Embeddings.